\documentclass[sigconf]{acmart}
\usepackage{bbding}
\usepackage{bm}
\usepackage[mathscr]{euscript}
\usepackage{subfigure}
\usepackage{tabularx,lipsum,environ,amsmath}
\usepackage{enumitem}
\usepackage[normalem]{ulem}



\newcommand{\specialcell}[2][c]{%
	\begin{tabular}[#1]{@{}c@{}}#2\end{tabular}}

\renewcommand\footnotetextcopyrightpermission[1]{} 
\AtBeginDocument{%
	\providecommand\BibTeX{{%
			\normalfont B\kern-0.5em{\scshape i\kern-0.25em b}\kern-0.8em\TeX}}}

\setcopyright{acmcopyright}
\copyrightyear{2018}
\acmYear{2018}
\acmDOI{10.1145/1122445.1122456}

\begin{document}

	\title{Knowledge-driven Site Selection via Urban Knowledge Graph}
	
%

 \author{Yu Liu}
 \affiliation{
   \institution{Tsinghua University}}
 \email{liuyu2419@126.com}

  \author{Jingtao Ding}
 \affiliation{
 	\institution{Tsinghua University}}
 \email{dingjt15@tsinghua.org.cn}

  \author{Yong Li}
  \affiliation{
    \institution{Tsinghua University}
  }
  \email{liyong07@tsinghua.edu.cn}
	
	\begin{abstract}
Site selection determines optimal locations for new stores, 
which is of crucial importance to business success. 
Especially, 
the wide application of artificial intelligence with multi-source urban data makes intelligent site selection promising. 
However, 
existing data-driven methods heavily rely on feature engineering, 
facing the issues of business generalization and complex relationship modeling. 
To get rid of the dilemma, 
in this work, 
we borrow ideas from knowledge graph (KG), 
and propose a knowledge-driven model for site selection, 
short for KnowSite. 
Specifically, 
motivated by distilled knowledge and rich semantics in KG, 
we firstly construct an urban KG (UrbanKG) with cities' key elements and semantic relationships captured. 
Based on UrbanKG, 
we employ pre-training techniques for semantic representations, 
which are fed into an encoder-decoder structure for site decisions. 
With multi-relational message passing and relation path-based attention mechanism developed, 
KnowSite successfully reveals the relationship between various businesses and site selection criteria. 
Extensive experiments on two datasets demonstrate that KnowSite outperforms representative baselines with both effectiveness and explainability achieved.	
	\end{abstract}
	
	
	
	
	\maketitle
	\section{Introduction}
	The task of site selection, 
	which selects optimal locations for opening new stores, is of crucial importance to business success. 
	A good choice of location always brings substantial profits while an inappropriate one could lead to store closure, such as opening a Starbucks store in a business area versus a residential one. 
	Generally, site selection for a specific brand requires a comprehensive consideration of both its own characteristics and those of potential urban regions, e.g., the brand's category and the region's human flow and function.  
	Traditional solution for most corporations is to employ expert consultants and conduct manual surveys \cite{phelps2018business,breheny1988practical,timmermans1986locational,kumar2000effect}, which are expensive, labor-intensive and time-consuming.

\begin{figure}[t]
		\centering
		\includegraphics[width=.85\linewidth]{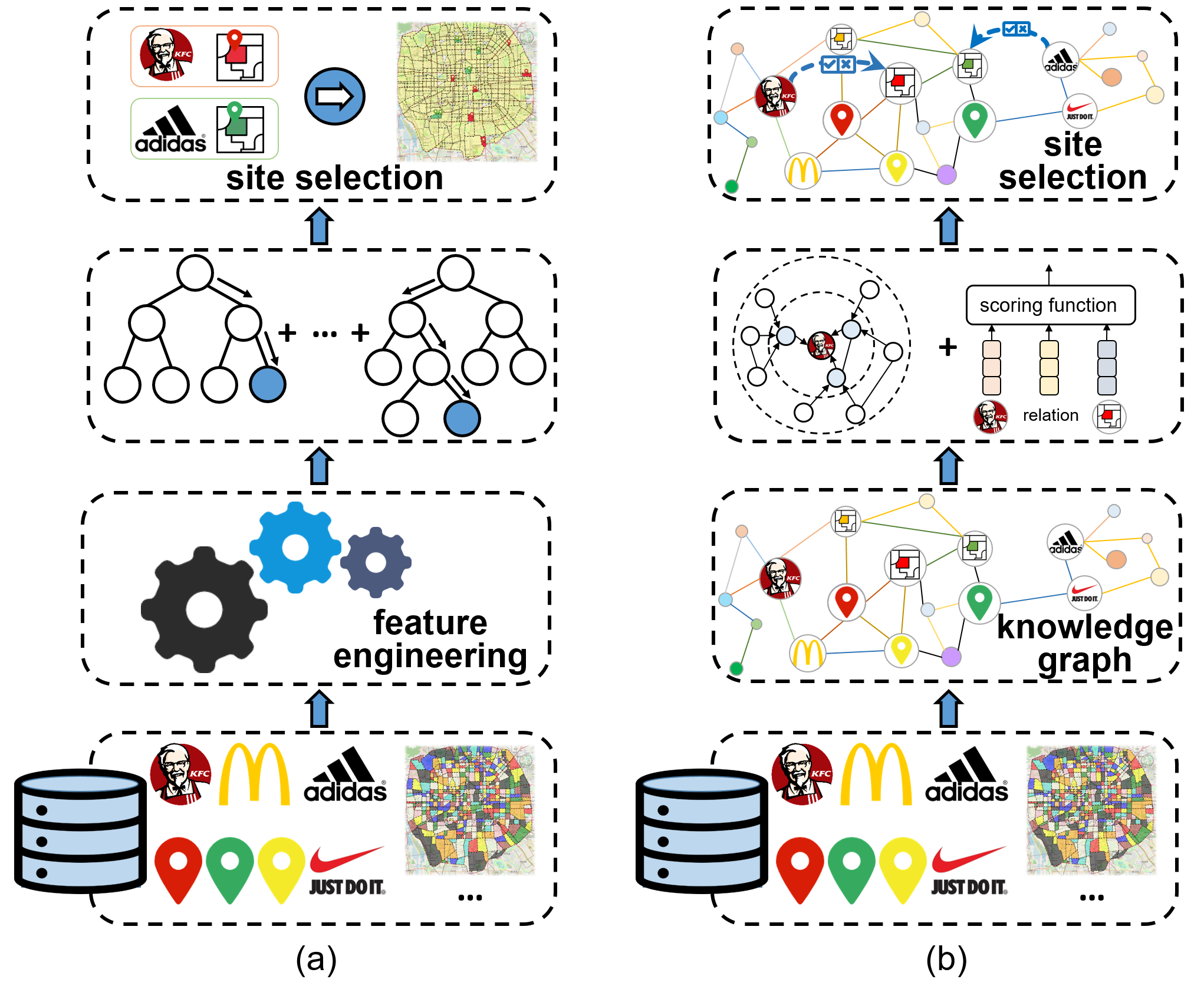}
		\vspace{-10px}
		\caption{Illustration of (a) data-driven paradigm and our proposed (b) knowledge-driven paradigm for site selection.}
		\label{fig:data-knowledge paradigm}
		\vspace{-10px}
	\end{figure}

	Owing to the rapid development of location-based services \cite{junglas2008location} and the wide availability of multi-source urban data \cite{zheng2014urban}, recent studies introduce the data-driven paradigm for site selection \cite{geo-spotting, liu2019deepstore, xu2020ar2net, xu2016demand, li2018commercial}. 
	As shown in Figure~\ref{fig:data-knowledge paradigm}(a), these data-driven approaches typically extract various features from the multi-source urban data, which are then fed into a machine learning model like XGBoost \cite{chen2016xgboost} to calculate the score for site decision. 
	However, 
	the manually defined feature involves only one or two aspects (store density, human flow, etc.), 
	failing to exploit complex relationships as well as diverse influences among multi-source urban data for site selection. 
	Moreover, such approaches merely provide an importance score for each feature without logical reasoning, 
	which is insufficient to persuade corporations \cite{timmermans1986locational,yap2018analytic}. 
	
	In comparison to the site selection still in data-driven paradigm, 
	several other areas of artificial intelligence have further introduced knowledge-driven paradigm for superior performance, 
	such as question answering \cite{huang2019knowledge}, natural language understanding \cite{zhang2019ernie} and recommender systems \cite{wang2019multi, guo2020survey}. 
	The core of such knowledge-driven paradigm is knowledge graph (KG) \cite{hogan2021knowledge}. 
	With domain entities as nodes and semantic relations as edges, 
	KG could integrate multi-source data into a graph structure, 
	and then powerful knowledge representation learning (KRL) methods are developed to avoid complex feature engineering \cite{ji2021survey}. 
	Hence, knowledge-driven paradigm stands out as a promising solution for site selection, but it is still underexplored due to following three critical challenges: 

	\begin{itemize}[leftmargin=*]
		\item \textbf{The difficulty of knowledge discovery from multi-source urban data.} Recently, the target knowledge for site selection lies in complex relationships among multi-source urban data, e.g., attribute, affiliation, spatiality, mobility and etc., which increases the difficulty to discover structured knowledge.
		\item \textbf{The complexity of knowledge refinement for diverse influences.} The influences of various knowledge are diverse for site selection, e.g., for KFC opening stores, the site decision of McDonald's is much more helpful than store density at regions. 		
		Thus, refining task-specific knowledge is non-trivial considering the rich while diverse urban contexts. 
		\item \textbf{The necessity of knowledge explainability to site decision understanding.} Although feature importance is provided in data-driven paradigm \cite{geo-spotting,xu2016demand}, the reasons for site decisions remains unknown, e.g., 
		finding new sites that have significant flow transition with existing sites. 
		So a challenge is how to clearly explain the logic behind corresponding site decisions for convincing and practical applications.   
	\end{itemize}
	
	To overcome the above challenges as well as explore the potential of KG, 
	in this paper, 
	we propose a generalized knowledge-driven paradigm for site selection. 
	As shown in Figure~\ref{fig:data-knowledge paradigm}(b), 
	we first construct the KG from multi-source urban data (referred to as UrbanKG), 
	based on which a generalized encoder-decoder structure is developed for site selection. 
	Specifically, knowledge discovery is achieved in UrbanKG, i.e.,
	the key elements of the city such as regions, point of interests (POIs), corporation brands, etc. are identified as entities, 
	while their complex relationships on attribute, affiliation, spatiality, mobility, etc. are modeled as relations. 
	To obtain semantic representations for entity and relation initialization, we adopt pre-training techniques on UrbanKG. 
	Furthermore, 
	we design a graph neural network (GNN) based encoder on UrbanKG, 
	such that knowledge refinement for diverse influences is adaptively modeled via multi-relational message passing. 
	As for the decoder part, 
	we carefully design a relation path based scoring function for knowledge explainability, 
	which measures the plausibility of site decisions between corporation brands and regions with the logical reasoning process revealed. 
	The scoring function firstly introduces multiple multi-hop relation paths based on different site selection criteria, 
	then generates relation path representations via semantic composition of relations, 
	and finally obtains corresponding scores using the attention mechanism. 
	The overall model is termed as KnowSite for \underline{Know}ledge-driven \underline{Site} selection. 
	Our key contributions are summarized as follows:
	\begin{itemize}[leftmargin=*]
		\item We are the first to propose the knowledge-driven paradigm for site selection, 
		and propose a model KnowSite generalized for various types of businesses. 
		Especially, KnowSite leverages urban knowledge via KG, 
		and builds an encoder-decoder structure to explore the knowledge for effective and explainable site selection. 
		\item We conduct a systematic study of knowledge discovery from multi-source urban data via KG construction, 
		which identifies key elements and complex relationships in the city as entities and relations, respectively. 
		\item Under the proposed encoder-decoder structure, 
		we design a multi-relational message passing mechanism with GNN based encoder for knowledge refinement, 
		and develop multi-hop relation path based decoder, 
		which achieves knowledge explainability with the reasons behind site decisions. 
		\item We conduct extensive experiments on two real-world datasets and the proposed KnowSite outperforms state-of-the-art data-driven  approaches by more than 18\% on precision, 
		which demonstrates the effectiveness of knowledge-driven paradigm. 
		Further visualization results shed light on understanding critical mechanism behind different brands' site decisions.
	\end{itemize}
	
	The rest of this paper is organized as follows. 
	Section~\ref{sec:problem} introduces the research problem, 
	while Section~\ref{sec:method} presents the details of our proposed knowledge-driven framework. 
	The empirical results are discussed in Section~\ref{sec:experiments}. 
	We review the related works in Section~\ref{sec:related_work}, 
	followed by a conclusion in Section~\ref{sec:conclusion}.

	\section{Problem Formulation}\label{sec:problem}
	Typically, the multi-source urban data for site selection can be categorized into three aspects \cite{guo2018citytransfer,geo-spotting,liu2019deepstore}. 
	
	\noindent\textbf{Spatial Data}. It includes the road network data $\mathcal{D}_{\textnormal{RN}}$ and business area (Ba) data $\mathcal{D}_{\textnormal{Ba}}$. $\mathcal{D}_{\textnormal{RN}}$ is a collection of road segments connected each other and $\mathcal{D}_{\textnormal{Ba}}$ collects core areas of business and commercial activities, e.g., \emph{Sanlitun}\footnote{\url{https://en.wikipedia.org/wiki/Sanlitun}} in Beijing, China.
	
	\noindent\textbf{Store Data}. It includes the POI data $\mathcal{D}_{\textnormal{POI}}$, brand data $\mathcal{D}_{\textnormal{Brand}}$ and site selection data $\mathcal{D}_{\textnormal{Site}}$. $\mathcal{D}_{\textnormal{POI}}$ and $\mathcal{D}_{\textnormal{Brand}}$ are the collection of venues and corporation brands respectively in the city. $\mathcal{D}_{\textnormal{Site}}$ records the brand and corresponding regions it opens a store. 
	
	\noindent\textbf{User Behavior Data}. It includes trajectory data $\mathcal{D}_{\textnormal{Traj}}$ with user trajectories, check-in data $\mathcal{D}_{\textnormal{Check}}$ with users' self-reported check-in records and click data $\mathcal{D}_{\textnormal{Click}}$ of aggregated clicking POIs records using map services.
	
	Then we formulate the knowledge-driven site selection problem. 
	\vspace{-15px}
	\newtheorem{problem}{Problem}
	\begin{problem}
		\textbf{\emph{Knowledge-driven Site Selection Problem.}}
		Given the multi source urban data, the knowledge-driven site selection problem can be divided into two sub-problems of KG construction and site selection. The KG construction sub-problem requires to construct KG $\mathcal{G}=f(\mathcal{D}_{\textnormal{RN}},\mathcal{D}_{\textnormal{Ba}},\mathcal{D}_{\textnormal{POI}},\mathcal{D}_{\textnormal{Brand}},\mathcal{D}_{\textnormal{Site}},\mathcal{D}_{\textnormal{Traj}},\mathcal{D}_{\textnormal{Check}},\mathcal{D}_{\textnormal{Click}})$ with construction method $f$. 
		Then the site selection sub-problem is formulated as a link prediction problem on $G$, predicting if there exists a site decision link between brand $b$ and region $a$, i.e., $(b,?,a)$.
	\end{problem}

	\section{Methodology} \label{sec:method}
	
	\subsection{Framework Overview}\label{sec:overview}
	\begin{figure}[htbp]
		\centering
		\vspace{-5px}
		\includegraphics[width=.85\linewidth]{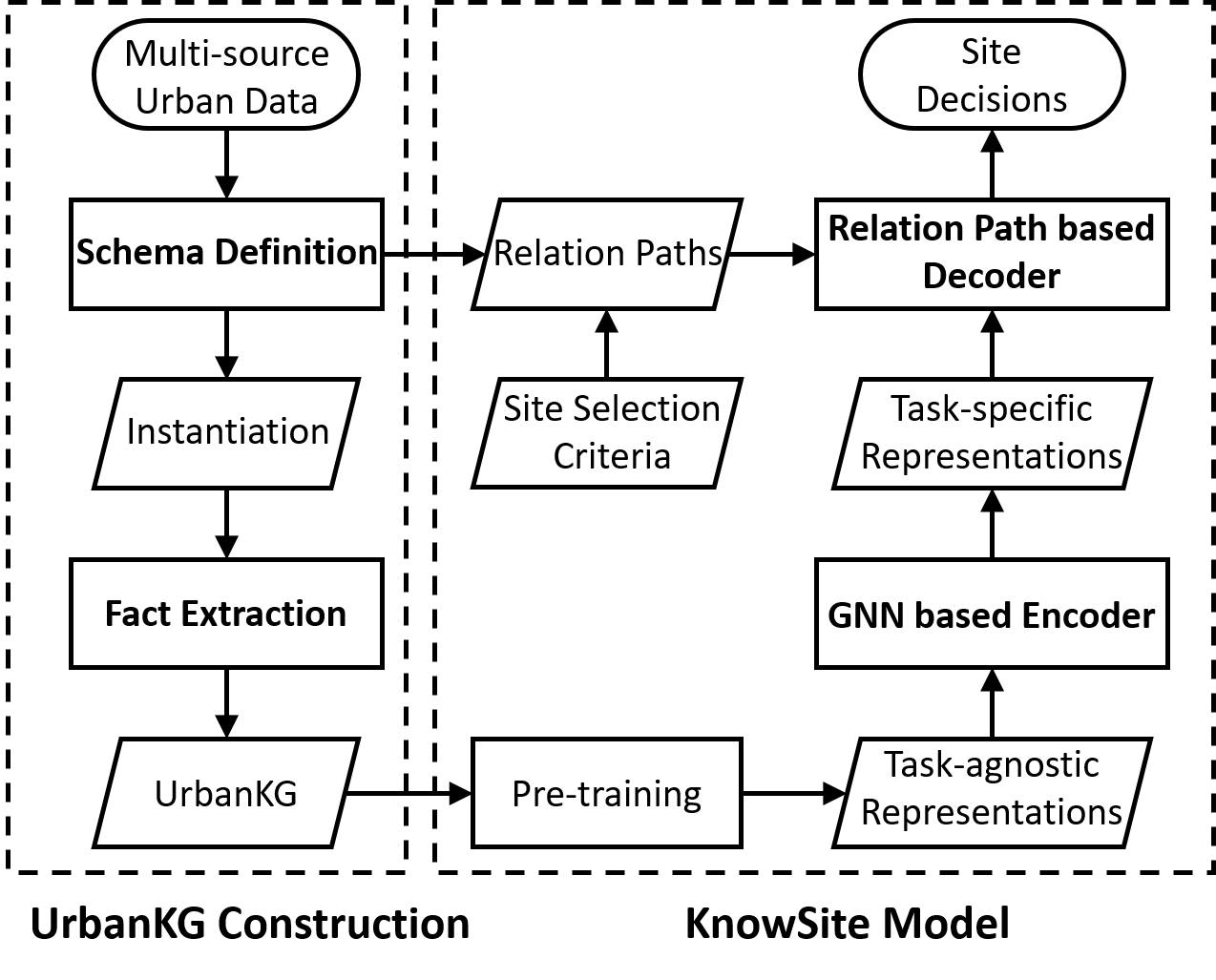}
		\vspace{-10px}
		\caption{The framework of our proposed knowledge-driven site selection method.}
		\label{fig:model_flow}
		\vspace{-10px}
	\end{figure}
	
	To overcome the challenges of applying knowledge-driven paradigm for site selection, 
	we present the framework of our proposed method in Figure~\ref{fig:model_flow}, 
	including UrbanKG construction and the KnowSite model for KG construction and site selection sub-problems, respectively. 
	Specifically, to discover knowledge from multi-source urban data, 
	we firstly construct UrbanKG for structured urban knowledge, 
	which is comprised of two major components: schema definition and fact extraction. 
	As for the KnowSite model, we develop the pre-training on UrbanKG for task-agnostic but knowledgeable representations. 
	To further refine knowledge for diverse influences, 
	we propose a GNN based encoder with task-specific representations learned. 
	Finally, to make knowledge explainable for identify reasons behind site decisions, 
	we design a relation path based decoder with effective performance achieved. 
	
	\subsection{UrbanKG Construction}\label{sec:urbankg_construction}
	To discover knowledge from multi-source urban data, 
	we construct UrbanKG for structured urban knowledge. 
	Formally, a KG is defined as a graph $\mathcal{G}=(\mathcal{E},\mathcal{R},\mathcal{F})$, 
	where $\mathcal{E}$ is the node set of entities and $\mathcal{R}$ is the edge set of relations, 
	while $\mathcal{F}$ corresponds to the fact set $\{(s,r,o)\mid s,o\in\mathcal{E},r\in\mathcal{R}\}$ \cite{wang2017knowledge,ji2021survey}. 
	The triplet $(s,r,o)$ denotes the directional edge from node $s$ to node $o$ via the edge of relation $r$. 
	\subsubsection{Schema Definition}\label{sec:schema definition}
\begin{figure}[t]
		\centering
		\includegraphics[width=.95\linewidth]{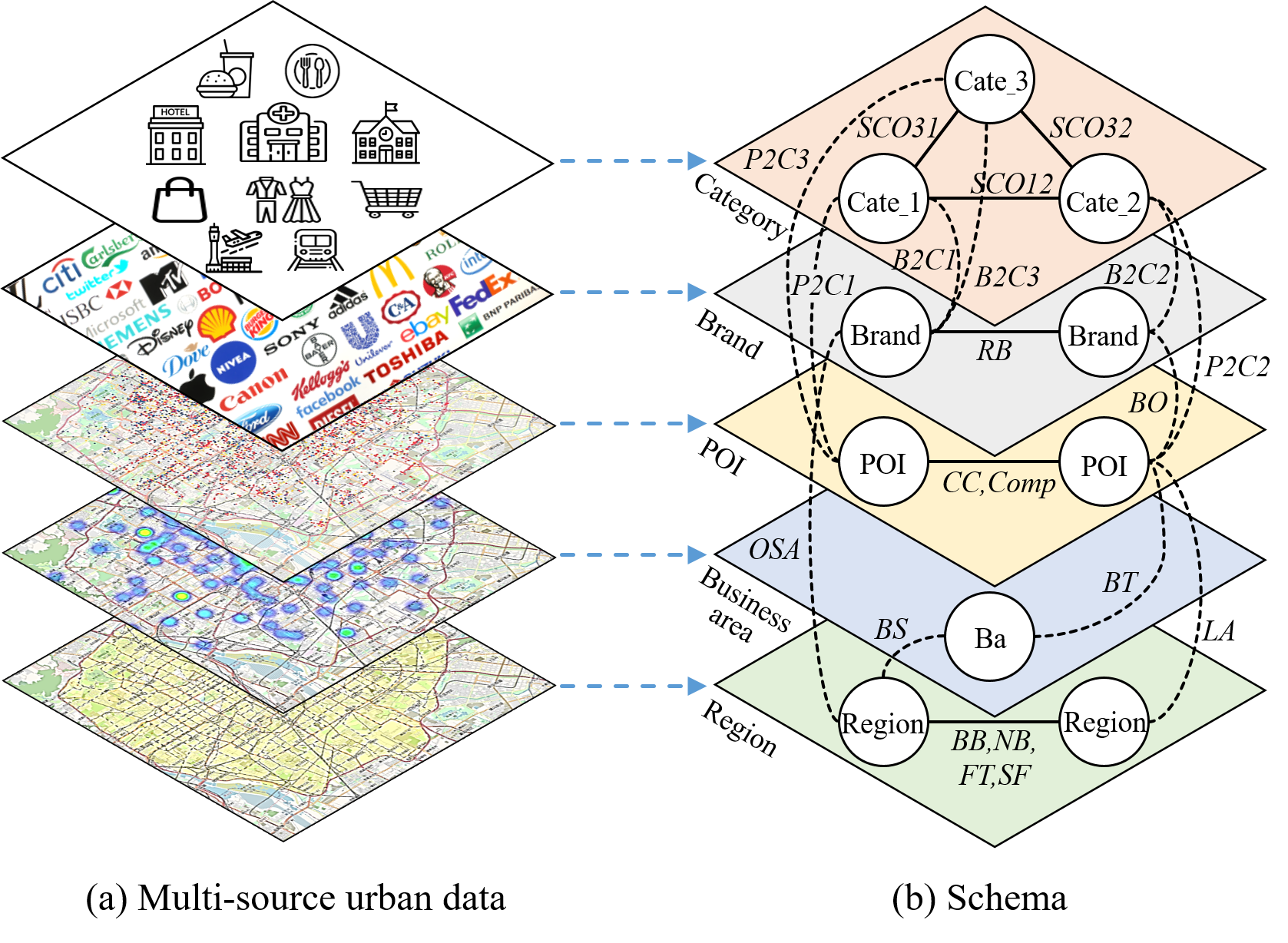}
		\vspace{-10px}
		\caption{The schema of urban knowledge graph. In (b), the dash lines represent inter-ontology relations and the solid lines for intra-ontology ones.}
		\label{fig:urbankg_construction}
		\vspace{-5px}
	\end{figure}

	At first, 
	by investigating the multi-source urban data, we build the schema of UrbanKG, as shown in Figure~\ref{fig:urbankg_construction}.  
	It defines a high-level structure for the KG with ontologies and relations \cite{hogan2021knowledge}, 
	where the ontologies determine the types of entities in UrbanKG, 
	including key elements in cities, i.e., Region, Ba, POI, Brand and Category, mainly identified from $\mathcal{D}_{\textnormal{RN}},\mathcal{D}_{\textnormal{Ba}},\mathcal{D}_{\textnormal{POI}}$ and $\mathcal{D}_{\textnormal{Brand}}$. 
	Since the category is an important property of POIs and brands, 
	we further divide the category into coarse-level, mid-level, and fine-grained categories, referred to as Cate\_1/2/3.

	Moreover, we identify the underlying relations to capture the complex relationships among city elements, 
	as presented in Table~\ref{tab:relations}. 
	For intra-ontology relations, 
	we describe them layer by layer, from bottom to up in Figure~\ref{fig:urbankg_construction}(b). 
	At the first layer of Region, 
	\emph{BorderBy} and \emph{NearBy} define the spatial relationships of two regions, 
	while \emph{SimilarFunction} link regions with similar POI distributios. 
	By analyzing $\mathcal{D}_{\textnormal{Traj}}$, we devise \emph{FlowTransition} to link regions with significant crowd flow transitions. 
	At POI layer, based on $\mathcal{D}_{\textnormal{Check}}$, 
	\emph{CoCheckin} reveals the geographical influence among POIs with check-in concurrence \cite{chang2020learning} and \emph{Competitive} models the competitive relationship among POIs \cite{li2020competitive}. 
	At Brand layer, 
	\emph{RelatedBrand} describes relatedness of brands. 
	At Category layer, 
	\emph{SubCateOf\_ij} defines the taxonomy among three-level categories.  
	As for inter-ontology relations, 
	\emph{BaServe}, \emph{BelongTo} and \emph{LocateAt} define the spatial relationships between different ontologies, especially \emph{BaServe} describes regions are in service range of business area. 
	Moreover, 
	\emph{POIToCate\_i} and \emph{BrandToCate\_i} represent the attribute relationships, 
	while \emph{BrandOf} describes the affiliation relationship between POI and brand. 
	\emph{OpenStoreAt} represents site selection records in $\mathcal{D}_\textnormal{Site}$.  
	Besides, for asymmetric relations $\{r\!\in\!\mathcal{R}\! \mid\! (s,r,o)\!\not\Leftrightarrow\!(o,r,s), \forall (s,r,o)\!\in\!\mathcal{F}\}$, 
	we introduce a new inverse relation $r^\prime$ into UrbanKG schema. 

	\begin{table}[htbp]
		\caption{The details of defined relations in UrbanKG.}\label{tab:relations}
		\vspace{-10px}
		\setlength\tabcolsep{0.8pt}
		\def\arraystretch{0.9}
		\begin{tabular}{c|c|c|c|c}
			\toprule
			Relation &  Abbrev. &  \specialcell{Subject \& Object\\ Ontologies} & Symmetry & Data Source \\
			\midrule
			\emph{BorderBy} &  \emph{BB} & (Region, Region)  & \Checkmark   & $\mathcal{D}_{\textnormal{RN}}$ \\
			\emph{NearBy} &  \emph{NB} & (Region, Region) & \Checkmark &  $\mathcal{D}_{\textnormal{RN}}$    \\
			\emph{FlowTransition} & \emph{FT} & (Region, Region) & \XSolidBrush & $\mathcal{D}_{\textnormal{Traj}}$ \\
			\emph{SimilarFunction} &  \emph{SF} & (Region, Region)  &\Checkmark & $\mathcal{D}_{\textnormal{POI}},\mathcal{D}_{\textnormal{RN}}$ \\
			\emph{CoCheckin}  &  \emph{CC} & (POI, POI) &    \Checkmark & $\mathcal{D}_{\textnormal{Check}},\mathcal{D}_{\textnormal{POI}}$     \\
			\emph{Competitive} &  \emph{Comp} & (POI, POI) &   \Checkmark &  $\mathcal{D}_{\textnormal{Brand}},\mathcal{D}_{\textnormal{POI}}$ \\
			\emph{RelatedBrand} & \emph{RB} & (Brand, Brand) & \Checkmark & $\mathcal{D}_{\textnormal{Brand}}$\\
			\emph{SubCateOf\_ij} &  \emph{SCOij} & (Cate\_\emph{i}, Cate\_\emph{j}) & \XSolidBrush  &$\mathcal{D}_{\textnormal{POI}}$  \\ 
			\emph{BaServe} &  \emph{BS} & (Ba, Region) &\XSolidBrush &  $\mathcal{D}_{\textnormal{Ba}},\mathcal{D}_{\textnormal{RN}}$  \\
			\emph{BelongTo} &  \emph{BT} & (POI, Ba) & \XSolidBrush &  $\mathcal{D}_{\textnormal{Ba}},\mathcal{D}_{\textnormal{POI}}$      \\
			\emph{LocateAt}  &  \emph{LA} & (POI, Region)  &   \XSolidBrush & $\mathcal{D}_{\textnormal{POI}},\mathcal{D}_{\textnormal{RN}}$    \\
			\emph{POIToCate\_i} &  \emph{P2Ci} & (POI, Cate\_\emph{i}) &   \XSolidBrush &   $\mathcal{D}_{\textnormal{POI}}$ \\
			\emph{BrandToCate\_i} &  \emph{B2C\_i} & (Brand, Cate\_\emph{i})  &   \XSolidBrush & $\mathcal{D}_{\textnormal{Brand}},\mathcal{D}_{\textnormal{POI}}$ \\
			\emph{BrandOf} &  \emph{BO} & (Brand, POI) & \XSolidBrush  & $\mathcal{D}_{\textnormal{Brand}},\mathcal{D}_{\textnormal{POI}}$ \\
			\emph{OpenStoreAt} &  \emph{OSA} & (Brand, Region) &  \XSolidBrush & $\mathcal{D}_{\textnormal{Site}}$  \\
			\bottomrule
		\end{tabular}
		\vspace{-10px}
	\end{table}
	
	\subsubsection{Fact Extraction}
	Based on the defined schema above, 
	we instantiate facts from the data, i.e., 
	mapping ontologies to specific entities and linking entities via semantic relations. 
	First, we introduce the mapping step. 
	For mapping Region ontology, we partition the city into disjointed regions according to the main road network with $\mathcal{D}_\textnormal{RN}$. 
	Compared with grid partition of equal size \cite{xu2020ar2net}, 
	our partition is much closer to people's movement and urban functional units \cite{shao2021deepflowgen}. 
	For Ba and POI ontologies, 
	we obtain the entities from $\mathcal{D}_\textnormal{Ba}$ and $\mathcal{D}_\textnormal{POI}$, respectively. 
	For Brand ontology, we adopt a text segmentation tool\footnote{\url{https://github.com/fxsjy/jieba}} and name matching to obtain entities. 
	For Category ontology, the three-level categories are divided by domain experts, 
	e.g., the Cate\_3 entity \emph{Beijing Cuisine} belongs to the Cate\_2 entity \emph{Chinese Food} and the Cate\_1 entity \emph{Food}. 
	Then, in the second step, the entities are further linked via relations defined in Table~\ref{tab:relations} with corresponding data sources. 
	Here we highlight the link details for brand-related relations. 
	For \emph{RelatedBrand}, the facts are obtained from a public KG \url{zhishi.me} with the “relatedPage” relation. 
	For \emph{BrandOf}, POI entities and their corresponding brand entities are linked together, based on which the \emph{BrandToCate\_i} facts are obtained by brands' connected POIs. 
	Other relational links follow the definitions above, and are obtained by data mapping, aggregation and calculation methods. 
	In this way, the constructed UrbanKG successfully presents the structured knowledge among multi-source urban data.
	
	\begin{figure}[htbp]
		\centering
		\includegraphics[width=.85\linewidth]{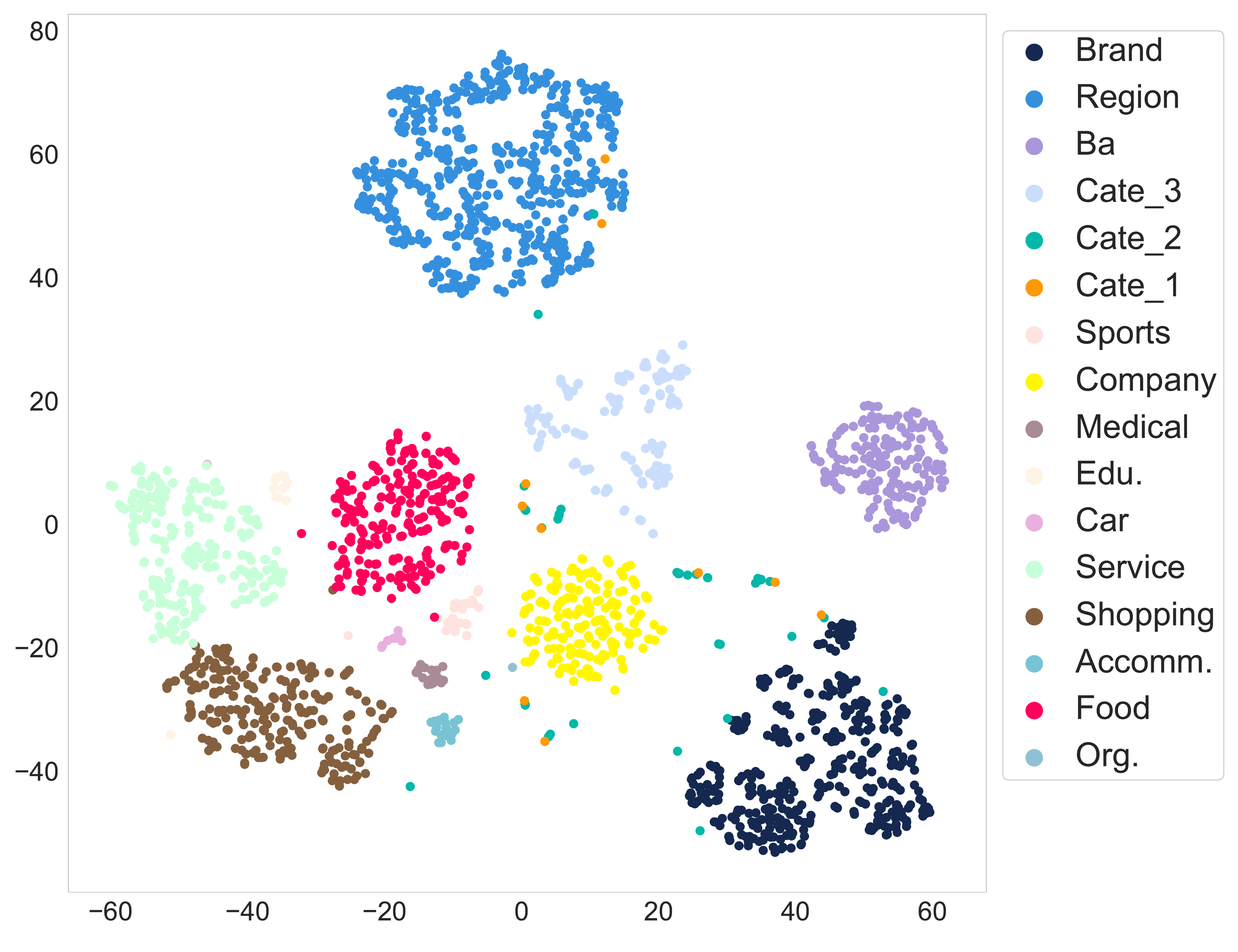}
		\vspace{-10px}
		\caption{t-SNE of pre-trained entity embeddings of beijing's UrbanKG (better viewed in color).}
		\label{fig:kg_vis}
		\vspace{-15px}
	\end{figure} 
	
	Using the data introduced in Section~\ref{sec:data} later, 
	we construct UrbanKGs for two of the largest cities, 
	which contains over 20k/40k entities and 300k/500k triplet facts in Beijing/Shanghai. 
	Note that the original UrbanKGs are significantly large, and we only report the statistics of subgraphs utilized in this work. 
	It is worth mentioning that we utilize pre-training for knowledgeable representations of entities and relations. 
	Specifically, 
	we leverage the KRL model, TuckER \cite{balazevic2019tucker} for pre-training, 
	which measures the plausibility of triplets in UrbanKG with embeddings learned. 
	Note that the pre-training process is task-agnostic and captures the global semantic information. 
	To validate the representation capability of UrbanKG, we visualize the
	pre-trained entity embeddings using t-SNE \cite{maaten2008visualizing}. 
	Especially, we randomly sample 1000 POI entities and all of other entities for visualization, as show in Figure~\ref{fig:kg_vis}. 
	It can be observed that, entities of the same ontology are clustered in space. 
	Moreover, POIs of different categories are also separated in visualization. 
	Such results indicate the effectiveness of our constructed UrbanKG with semantics captured.

	\subsection{The KnowSite Model}\label{sec:model}
	\subsubsection{GNN based Encoder}\label{sec:encoder}
	To fully explore the potential of UrbanKG and model diverse influences of various knowledge, 
	we design the GNN based encoder for knowledge refinement. 
	
	For a node/entity $v$ in KG $\mathcal{G}=(\mathcal{E},\mathcal{R},\mathcal{F})$, 
	$d$ denotes the embedding dimension, 
	$\bm{h}^{k}_v\in\mathbb{R}^d$ denotes its representation after $k$ layers GNN, 
	while $\mathcal{N}^r_v$ denotes its neighbors under relation $r\in\mathcal{R}$. 
	The relation $r$'s representation at layer $k$ is denoted as $\bm{h}_r^k\in\mathbb{R}^d$. 
	The number of GNN layers is denoted as $K$. 
	Especially, 
	the representation of node $v$ at layer $k+1$, $\bm{h}_v^{k+1}$ is obtained via three steps \cite{schlichtkrull2018modeling,vashishth2019composition,wang2021relational,ying2019gnnexplainer,yu2021knowledge}. 
	(1) Message calculation, which defines the function \textit{MSG} to calculate the message for triplet $(u,r,v)$: $m^{k+1}_{urv}=\textit{MSG}(\bm{h}^k_u,\bm{h}^k_r,\bm{h}^k_v)$.
	(2) Message aggregation, which defines the function \textit{AGG} to aggregate messages from node $v$'s neighbors: 
	$M^{k+1}_v=\textit{AGG}(m^{k+1}_{urv}|r\!\in\!\mathcal{R},u\!\in\!\mathcal{N}_v^r)$.
	(3) Representation update, which defines the function \textit{UPD} to update $v$'s representation from the aggregated messages $M^{k+1}_v$ and $v$'s previous layer representation $\bm{h}^k_v$: $\bm{h}^{k+1}_v=UPD(\bm{h}^{k}_v, M^{k+1}_v)$. 
	
	In terms of message calculation, 
	for a node $v$ with the triplet $(u,r,v)$, 
	our proposed GNN based encoder adopts the composition of neighbor node and linked relation \cite{vashishth2019composition,schlichtkrull2018modeling}:
	\setlength{\abovedisplayskip}{1pt}
	\setlength{\belowdisplayskip}{1pt}
	\begin{align}
		\setlength{\abovedisplayskip}{0pt}
		\setlength{\belowdisplayskip}{0pt}
		\textit{MSG}\left(\bm{h}^k_u,\bm{h}^k_r,\bm{h}^k_v\right) = \bm{W}^k_r\phi\left(\bm{h}^k_u,\bm{h}^k_r\right),
	\end{align}
	where $\bm{W}^k_r$ is the relation-specific projection matrix, 
	while $\phi:\mathbb{R}^d\times\mathbb{R}^d\rightarrow\mathbb{R}^d$ is the entity-relation composition operation, e.g., element-wise subtraction and element-wise multiplication. 
	
\begin{figure}[t]
		\centering
		\includegraphics[width=.99\linewidth]{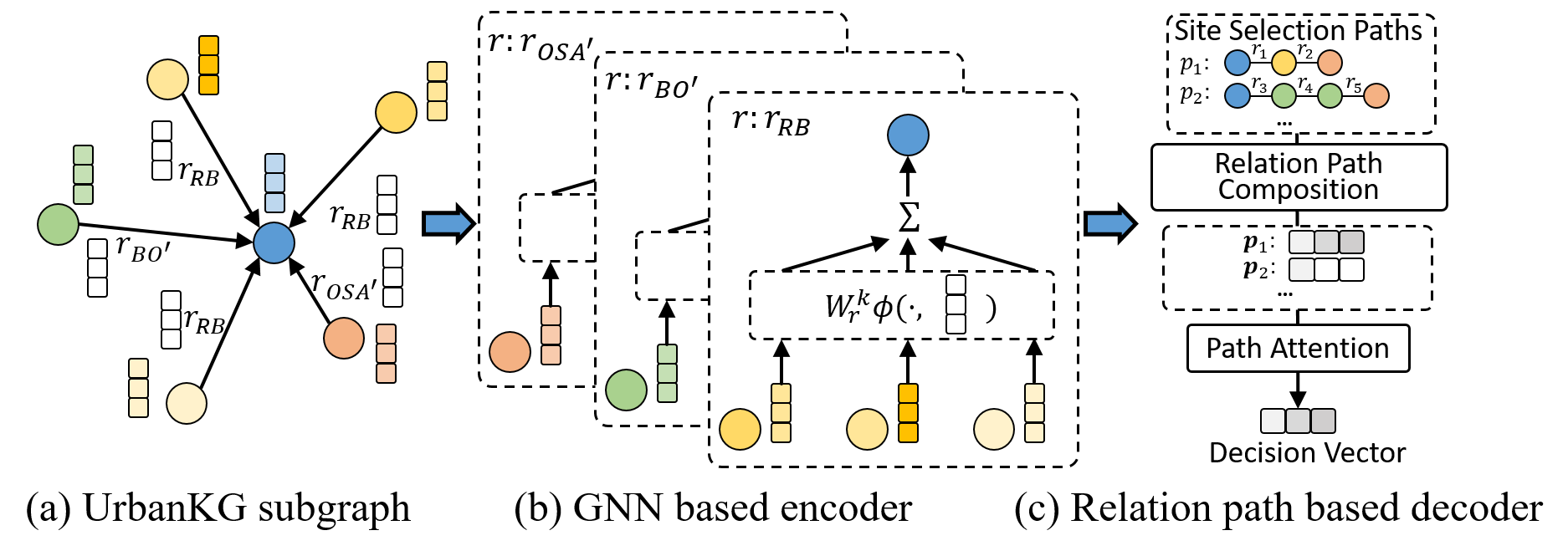}
		\vspace{-12px}
		\caption{The illustration of KnowSite model with a subgraph of UrbanKG.}
		\label{fig:knowsite}
		\vspace{-15px}
	\end{figure}

	Moreover, the message aggregation and the representation update are defined as relation-specific mean pooling and nonlinear transformation, respectively. 
	Thus, 
	the representation of node $v$ at layer $k+1$ can be expressed as, 
	
	\begin{align}
		\bm{h}^{k+1}_{v}=f\left( \sum_{r\in\mathcal{R}}\frac{1}{\left|\mathcal{N}^r_v\right|} \sum_{u\in\mathcal{N}^r_v} \bm{W}^k_r \phi\left(\bm{h}^k_u, \bm{h}^k_r\right) \right), 
	\end{align}
	where $\mathcal{N}^r_v$ denotes $v$'s neighbors under relation $r\!\in\!\mathcal{R}$, 
	while $f\!:\!\mathbb{R}^d\!\rightarrow\!\mathbb{R}^d$ denotes the nonlinear activation function. 
	Such relation-specific message passing is illustrated from Figure~\ref{fig:knowsite}(a) to (b). 
	Besides, 
	in each layer the relation representation is obtained via linear projection, 
	\begin{align}
		\bm{h}_r^{k+1}=\bm{W}^{k+1}_{\textnormal{rel}}\bm{h}_r^k,
	\end{align}
	where $\bm{W}^{k+1}_{\textnormal{rel}}$ denotes the relational projection matrix at layer $k+1$. 
	The pre-trained embeddings are initialized for $\bm{h}_r^0,\bm{h}_{u}^k,\bm{h}_{v}^k$.
	
	Compared with task-agnostic pre-training, 
	the GNN based encoding is task-specific, 
	where the learnable projection matrices determine the influences of various messages and refine useful knowledge for site selection, 
	supervised by the task loss introduced in the following. 
	Such multi-relational message passing mechanism

	\subsubsection{Relation Path based Decoder}\label{sec:decoder}
	With knowledgeable representations obtained in GNN based encoder, 
	to explore the explainability of knowledge, 
	we further propose a relation path based decoder for both effective and explainable site decisions. 
	Here we first introduce the relation path in KG \cite{lin2015modeling,zhu2019representation}.
	\begin{definition}\label{def:relation_path}
		\textbf{Relation Path.} 
		A relation path in KG $\mathcal{G}=(\mathcal{E},\mathcal{R},\mathcal{F})$ is defined as $p=(r_1,\cdots,r_{\left|p\right|})$, 
		where $\left|p\right|$ denotes the number of hops and $r_1,\cdots,r_{\left|p\right|}\in\mathcal{R}$. 	
	\end{definition}
	
	Obviously, the relation path provides rich semantic contexts and can be used to explain the logical reasoning of site decisions with UrbanKG. 
	For example, 
	the relation path
	$\textnormal{Brand}\stackrel{r_{OSA}}{\longrightarrow}\textnormal{Region}\stackrel{r_{SF}}{\longrightarrow}\textnormal{Region}$ focuses on the criteria of region function, 
	i.e., opening the new store at the region with similar functions, 
	while $\textnormal{Brand}\stackrel{r_{RB}}{\longrightarrow}\textnormal{Brand}\stackrel{r_{OSA}}{\longrightarrow}\textnormal{Region}$ indicates the logic that the brand learns from its related brand and opens the new store at the same region. 
	Thus, 
	based on UrbanKG and key criteria in traditional site selection \cite{yildiz2019hybrid,yap2018analytic}, 
	we summarize the relation paths for site selection in Table~\ref{tab:relation_path}. 
	
	\begin{table}[htbp]
		\caption{Relation paths for site selection in UrbanKG. Relations of $r_{\textnormal{BS\_1}}^\prime$, $r_{\textnormal{B2C\_1}}^\prime$ $r_{\textnormal{P2C\_1}}^\prime$ represent inverse relations.}\label{tab:relation_path}
		\vspace{-10px}
		
		\def\arraystretch{0.9}
		\begin{tabular}{c|l}
			\toprule
			Criteria            & Relation Paths with Ontologies \\
			\midrule
			Region Distance            &      $\textnormal{Brand}\stackrel{r_{OSA}}{\longrightarrow}\textnormal{Region}\stackrel{r_{NB}}{\longrightarrow}\textnormal{Region}$      \\
			Region Function            &  $\textnormal{Brand}\stackrel{r_{OSA}}{\longrightarrow}\textnormal{Region}\stackrel{r_{SF}}{\longrightarrow}\textnormal{Region}$     \\
			Region Flow                &       $\textnormal{Brand}\stackrel{r_{OSA}}{\longrightarrow}\textnormal{Region}\stackrel{r_{FT}}{\longrightarrow}\textnormal{Region}$   \\
			Business Area       &       $\textnormal{Brand}\stackrel{r_{OSA}}{\longrightarrow}\textnormal{Region}\stackrel{r_{BS}^\prime}{\longrightarrow}\textnormal{Ba}\stackrel{r_{BS}}{\longrightarrow}\textnormal{Region}$  \\
			Related Brand       &     $\textnormal{Brand}\stackrel{r_{RB}}{\longrightarrow}\textnormal{Brand}\stackrel{r_{OSA}}{\longrightarrow}\textnormal{Region}$                   \\
			Brand Category &      $\textnormal{Brand}\stackrel{r_{B2C\_1}}{\longrightarrow}\textnormal{Cate\_1}\stackrel{r_{{B2C\_1}}^\prime}{\longrightarrow}\textnormal{Brand}\stackrel{r_{OSA}}{\longrightarrow}\textnormal{Region}$   \\
			Competitiveness     &       $\textnormal{Brand}\stackrel{r_{BO}}{\longrightarrow}\textnormal{POI}\stackrel{r_{Comp}}{\longrightarrow}\textnormal{POI}\stackrel{r_{LA}}{\longrightarrow}\textnormal{Region}$     \\
			Store Category &     $\textnormal{Brand}\stackrel{r_{B2C}}{\longrightarrow}\textnormal{Cate\_1}\stackrel{r_{{P2C\_1}}^\prime}{\longrightarrow}\textnormal{POI}\stackrel{r_{LA}}{\longrightarrow}\textnormal{Region}$   \\
			\bottomrule
		\end{tabular}
		\vspace{-15px}
	\end{table}
	
	Based on the relation paths, 
	we introduce the design of relation path based decoder, 
	as shown in Figure~\ref{fig:knowsite}(c). 
	First, 
	we obtain the representation of each relation path by semantic composition \cite{lin2015modeling}. 
	Specifically, 
	given a relation path $p=(r_1,\cdots,r_{\left|p\right|})$ and the brand $b$, 
	the brand-specific path representation can be calculated via following three ways, 
	\begin{align}
		\textnormal{Add}:
		&	\bm{p} = \bm{h}^K_b + \bm{h}^K_{r_1} + \cdots + \bm{h}^K_{r_{\left|p\right|}},\\
		\textnormal{Mult}:
		&\bm{p} = \bm{h}^K_b \odot \bm{h}^K_{r_1} \odot \cdots \odot \bm{h}^K_{r_{\left|p\right|}},\\
		\textnormal{GRU}: 
		&	\bm{p} = \textnormal{GRU}([\bm{h}^K_{r_1}, \cdots, \bm{h}^K_{r_{\left|p\right|}}], \bm{h}^K_b), \label{eq:GRU}
	\end{align}
	where $\odot$ is the element-wise product, 
	and $\bm{h}^K_b$ in \eqref{eq:GRU} is the initial hidden state for gated recurrent unit (GRU) input.
	
	Since multiple factors/criteria are comprehensively considered in site selection \cite{turhan2013literature}, 
	we further applies the attention mechanism \cite{vashishth2019composition} on relation paths for brand-specific site decision vector, 
	\begin{align}
		\bm{z}_b = \textnormal{Attention}(\bm{W}^{\textnormal{Query}}\bm{h}^K_b,\bm{W}^{\textnormal{Key}}\bm{P}, \bm{W}^{\textnormal{Value}}\bm{P}),\label{eq:attn}
	\end{align}
	where $\bm{P}=[\bm{p}_1;\cdots;\bm{p}_{n_p}]$ is the concatenated relation path representation matrix and $n_p$ is the number of relation paths for site selection ($n_p=8$ in our case). 
	$\bm{W}^{\textnormal{Query}},\bm{W}^{\textnormal{Key}}$ and $\bm{W}^{\textnormal{Value}}$ are learnable parameters in the attention mechanism. 
	The attention weights provide explainable results behind site decisions, especially the relationship between brands and various criteria.
	
	For pairwise data $(b,a)\in\mathcal{D}_\textnormal{site}$ ($b$ is the brand and $a$ is the region), 
	the decision vector is multiplied with region embedding vector for the path based score. 
	Additionally, 
	for relatedness maximization, we utilize the bilinear product\footnote{$\langle\bm{a},\bm{b},\bm{c}\rangle=\sum_i a_i\cdot b_i \cdot c_i$} to obtain the link based score via direct relation $OpenStoreAt$. 
	The two parts are fused by a hyper-parameter $\alpha$ for final link prediction score on site selection.
	\begin{align}
		y_{ba} = (1-\alpha)\cdot\bm{z}_b^\top\bm{h}^K_a + \alpha\cdot \langle \bm{h}^K_b,\bm{h}^K_{r_{OSA}},\bm{h}^K_a \rangle. \label{eq:score}
	\end{align}
	
	Accordingly, 
	we adopt the cross-entropy loss for model parameter learning, 
	and formulate the objective function as follow, 
	\begin{align}
		\min\limits_{\bm{\Theta}} \sum_{(b_i,a_j)\in\mathcal{D}_\textnormal{Site}} -\log \frac{e^{y_{b_ia_j}}}{\sum_{a_k\in\mathcal{A}}e^{y_{b_ia_k}}} + \lambda\cdot\|\bm{\Theta}\|, \label{eq:loss}
	\end{align}
	where $\bm{\Theta}$ includes the learnable parameters in GNN based encoder and relation path based decoder. 
	$\mathcal{A}$ represents the set of candidate regions. 
	$\lambda$ is used to regularize the model parameters.
	The proposed KnowSite model is trained in a mini-batch way to minimize
	the objective formulation above. 
	
	Overall, 
	with task-specific loss and end-to-end training, 
	the proposed KnowSite model designs the multi-relational GNN based encoder for site selection related message passing, 
	and further learns the relation path based decoder to explicitly model the logic of site decisions, 
	achieving both effective and explainable performance. 
	
	\section{Evaluation}\label{sec:experiments}
	
	\subsection{Experimental Setup}
	\subsubsection{Datasets}\label{sec:data}
	We collect two datasets for evaluation with two cities of Beijing and Shanghai. 
	Several sources of urban data are collected and crawled from map service, life service platform, social media\footnote{\url{https://weibo.com}, \url{https://www.amap.com},  \url{https://www.meituan.com}} as well as Internet service provider, 
	which are summarized in Table~\ref{tab:data}. 
	Besides, the user data has been anonymized for privacy protection. 
	The details of site selection data and UrbanKG can be found in Section~\ref{app:data_details}.
	
	\begin{table}[htbp]
		\caption{A summary of the multi-source urban data collected for Beijing/Shanghai in China. ISP denotes Internet Service Provider. $n$ denotes the sequence length.}\label{tab:data}
		\vspace{-10px}
		\setlength\tabcolsep{0.pt}
		
		\begin{tabular}{cccc}
			\toprule
			Data     & Format  & Source & Amount \\
			\midrule
			$\mathcal{D}_{\textnormal{RN}}$  & $(\text{region\_id}, \{(\text{lng}_i,\text{lat}_i)\}_{i=1}^n)$ & Map Service  & 500/2k  \\
			$\mathcal{D}_{\textnormal{Ba}}$ & $(\text{Ba\_id}, \text{name}, \text{lng},\text{lat})$ & Life Service Platform & 160/200	\\
			$\mathcal{D}_{\textnormal{POI}}$ & $(\text{pid}, \text{name}, \text{lng}, \text{lat}, \text{cate}\_{1/2/3})$ & Map Service & 22k/38k\\
			$\mathcal{D}_{\textnormal{Brand}}$ & $(\text{brand\_id},\text{name})$ & Text Mining &  400 \\
			$\mathcal{D}_{\textnormal{Site}}$ & $(\text{brand\_id},\text{region\_id})$ &$\mathcal{D}_{\textnormal{POI}},\mathcal{D}_{\text{Brand}},\mathcal{D}_{\textnormal{RN}}$  & 25k/40k\\
			$\mathcal{D}_{\textnormal{Traj}}$ & $(\text{uid},\{(\text{lng}_i,\text{lat}_i,t_i)\}_{i=1}^n)$ & ISP & 400k/150k\\
			$\mathcal{D}_{\textnormal{Check}}$ &  $(\text{uid},\text{pid}, \#\text{check-in})$ & Social Media & 1M \\
			$\mathcal{D}_{\textnormal{Click}}$ & $(\text{pid}, \#\text{click})$ & Map Service &  22k/38k\\
			\bottomrule     
		\end{tabular}
		\vspace{-10px}
	\end{table}

	\begin{table*}[htbp]
		\caption{Performance comparison w.r.t. test NDCG@k, Hit@k, Precision@k, Recall@k and MAP@k on two datasets. Best results are in bold and the best results (in baselines) are underlined. The last two rows show relative improvement in percentage and $p$-value compared with the best baseline with 10 runs of experiments.}\label{tab:main_results}
		\vspace{-10px}
		\setlength\tabcolsep{3.5pt}
		\def\arraystretch{0.95}
		\begin{tabular}{cccccccc|ccccccc}
			\toprule
			& \multicolumn{7}{c}{Beijing}                           & \multicolumn{7}{c}{Shanghai}                          \\
			\midrule
			Model    & N@5   & N@10  & H@5   & H@10  & P@10  & R@10  & M@10  & N@5   & N@10  & H@5   & H@10  & P@10  & R@10  & M@10  \\
			\midrule
			Lasso    & 0.057 & 0.061 & 0.189 & 0.305 & 0.061 & 0.068 & 0.031 & 0.039 & 0.037 & 0.118 & 0.176 & 0.037 & 0.038 & 0.020 \\
			SVR      & 0.094 & 0.093 & 0.301 & 0.435 & 0.082 & 0.096 & 0.046 & 0.064 & 0.059 & 0.211 & 0.299 & 0.054 & 0.058 & 0.028 \\
			XGBoost  & 0.100 & 0.100 & 0.320 & 0.454 & 0.089 & 0.103 & 0.050 & 0.075 & 0.062 & 0.205 & 0.297 & 0.058 & 0.062 & 0.030 \\
			RankNet  & 0.122 & 0.121 & 0.369 & 0.501 & 0.104 & 0.122 & 0.064 & 0.085 & 0.081 & 0.274 & 0.383 & 0.074 & 0.078 & 0.038 \\
			NeuMF-RS & 0.180 & 0.178 & 0.501 & 0.653 & 0.155 & 0.182 & 0.097 & 0.178 & 0.168 & 0.478 & 0.615 & 0.148 & 0.163 & 0.090 \\
			\midrule
			TransE   & 0.080 & 0.084 & 0.297 & 0.460 & 0.075 & 0.089 & 0.036 & 0.064 & 0.063 & 0.244 & 0.372 & 0.058 & 0.064 & 0.026 \\
			DistMult & 0.161 & 0.161 & 0.475 & 0.634 & 0.137 & 0.164 & 0.083 & 0.150 & 0.142 & 0.448 & 0.591 & 0.124 & 0.138 & 0.071 \\
			ComplEx  & 0.170 & 0.169 & 0.502 & 0.657 & 0.143 & 0.171 & 0.088 & 0.147 & 0.142 & 0.442 & 0.583 & 0.126 & 0.140 & 0.070 \\
			TuckER   & \underline{0.183} & \underline{0.183} & \underline{0.518} & \underline{0.673} & \underline{0.156} & \underline{0.187} & \underline{0.098} & \underline{0.188} & \underline{0.174} & \underline{0.502} & \underline{0.620} & \underline{0.150} & \underline{0.166} & \underline{0.094} \\
			\midrule
			KnowSite (Add) & {0.218} & {0.217} & {0.556} & {0.707} & {0.185} & {0.222} & {0.125} & {0.218} & {0.200} & {0.541} & {0.653} & {0.171} & {0.191} & {0.113} \\
			KnowSite (Mult) & \textbf{0.221} & \textbf{0.219} & \textbf{0.565} & {0.709} & \textbf{0.186} & \textbf{0.224} & \textbf{0.127} & {0.219} & {0.202} & \textbf{0.543} & {0.664} & {0.173} & {0.193} & {0.115} \\
			KnowSite (GRU) & {0.220} & \textbf{0.219} & {0.557} & \textbf{0.713} & \textbf{0.186} & {0.223} & \textbf{0.127} & \textbf{0.220} & \textbf{0.205} & \textbf{0.543} & \textbf{0.671} & \textbf{0.177} & \textbf{0.197} & \textbf{0.116} \\
			\hline\hline
			Improv. & \textbf{20.8\%} & \textbf{19.7\%} & \textbf{9.1\%} & \textbf{5.9\%} & \textbf{19.2\%} & \textbf{19.8\%} & \textbf{29.6\%} & \textbf{17.0\%} & \textbf{17.8\%} & \textbf{8.2\%} & \textbf{8.2\%} & \textbf{18.0\%} & \textbf{18.7\%} & \textbf{23.4\%} \\
			$p-$value & {2.0e-10} & {1.5e-11} & {1.8e-6} & {1.6e-5} & {7.1e-12} & {1.1e-11} & {1.5e-11} & {1.1e-9} & {4.2e-11} & {6.9e-8} & {3.0e-10} & {1.9e-11} & {6.6e-11} & {1.2e-9} \\
			\bottomrule
		\end{tabular}
		\vspace{-10px}
	\end{table*}

	\subsubsection{Baselines}
	We compare our proposed KnowSite model with two types of models. 
	First, 
	following the feature engineering and framework in \cite{geo-spotting,xu2016demand,xu2020ar2net,liu2019deepstore,li2018commercial}, 
	we choose five traditional data-driven models, 
	Lasso \cite{lasso}, SVR \cite{svr}, XGBoost \cite{chen2016xgboost}, RankNet \cite{ranknet} as well as NeuMF-RS \cite{li2018commercial}. 
	All data sources have been utilized for feature extraction. 
	Due to the model generalization issue to various brands, 
	we train and test the first four models brand by brand, 
	and report the average performance. 
	Second, 
	we further compare with four typical KG link prediction models on UrbanKG, 
	TransE \cite{bordes2013translating}, DistMult \cite{yang2014embedding}, ComplEx \cite{trouillon2016complex} and TuckER \cite{balazevic2019tucker}. 
	All the baselines are tuned with their reported settings (in site selection works, if applicable), 
	and the weights of $OpenStoreAt$ links in KG completion models are increased to 10 for the site selection task.

	\subsubsection{Evaluation Metrics}
	We evaluate the site selection performance with five standard metrics of $\textnormal{NDCG}@k$, $\textnormal{Hit}@k$, $\textnormal{Precision}@k$, $\textnormal{Recall}@k$ and $\textnormal{MAP}@k$  \cite{geo-spotting,xu2020ar2net,xu2016demand,li2018commercial} that defined in Section~\ref{app:metrics}. 
	We evaluate the performance with $k=5,10,20$. 
	Due to the space limitation, some results with $k=5,20$ are omitted, 
	which are in accord with other metrics.
	
	\subsubsection{Implementation}
	For the proposed KnowSite model learning, 
	the batch size is set to 128 and the embedding dimension $d$ is set to 64. 
	Besides, batch normalization and dropout are used for regularization. 
	We use the rotate composition operator \cite{galkin2020message} in GNN based encoder, 
	and the number of GNN layers ranges from 1 to 3. 
	We tune other hyper-parameters with early stopping mechanism on validation $\textnormal{NDCG}@10$. 
	The learning rate and the dropout are searched from $\{0.0005, 0.001, 0.003, 0.005\}$ and $\{0.1, 0.3, 0.5\}$, respectively. 
	The fusion parameter $\alpha$ ranges from $0.0$ to $1.0$. 
	As for the pre-training step, 
	we train the TuckER model with early stopping mechanism on training loss. 
	All models are run 10 times and the average results are reported to prevent extreme cases. 
	Besides, the stores (POIs) as well as $OpenStoreAt$ links in valid \& test sets are removed from UrbanKG to avoid test leakage. 
	
	Next, 
	we present the performance comparison on two datasets, and then analyze the effectiveness of each module in KnowSite with ablation study. 
	Several explainable results are further investigated for the logic of site selection.
	
	\subsection{Performance Comparison}
	Table~\ref{tab:main_results} presents the site selection performance comparison on both datasets. 
	For KnowSite, 
	all three composition operations of addition (Add), multiplication (Mult) and GRU are considered for relation path representation. 
	In general, our proposed KnowSite outperforms all baselines across five evaluation metrics. 
	Specifically, 
	the improvement in Beijing dataset ranges from 5.9\%\textasciitilde29.6\%, 
	while the improvement in Shanghai dataset is from 8.2\%\textasciitilde23.4\%. 
	The considerable improvements demonstrate the effectiveness of our proposed knowledge-driven paradigm as well as systematic encoder-decoder framework. 
	Besides, 
	KnowSite models with three composition operations achieve comparable performance, 
	and we select the GRU operation for detailed studies later. 
	
	Moreover, we have following three observations. 
	First, knowledge-driven models of DistMult, ComplEx, TuckER and KnowSite perform more competitively than left data-driven ones, which owes to the knowledge discovery on UrbanKG. 
	For example, the best data-driven baseline NeuMF-RS formulates the problem as matrix completion, 
	which is easily affected by limited brand-region samples and cannot exploit rich semantics in multi-source urban data as UrbanKG does. 
	Second, knowledge-driven models show strong robustness to various cities with knowledge refinement. 
	For the two datasets, Shanghai dataset contains much more brands and candidate regions, and thus is more challenging. 
	Due to the incompleteness of feature engineering and diverse influences, the performance gap of data-driven models between the two datasets are significant, e.g., a gap of over 0.150 on Hit@10 for SVR/XGBoost. 
	In comparison, the gap for knowledge-driven models is less than 0.080 with site selection knowledge learned. 
	Third, the performance gap between KG link prediction models and KnowSite implies that extending KRL methods to site selection application is nontrivial and needs further customized designs, e.g., multi-relational message passing for knowledge refinement, and site selection related relation paths as well as brand-specific attention mechanism for knowledge explainability.
	
	\subsection{Ablation Study}
\begin{figure}[hbtp]
		\vspace{-10px}
		\centering
		\subfigure[Beijing]{
			\includegraphics[width=0.231\textwidth]{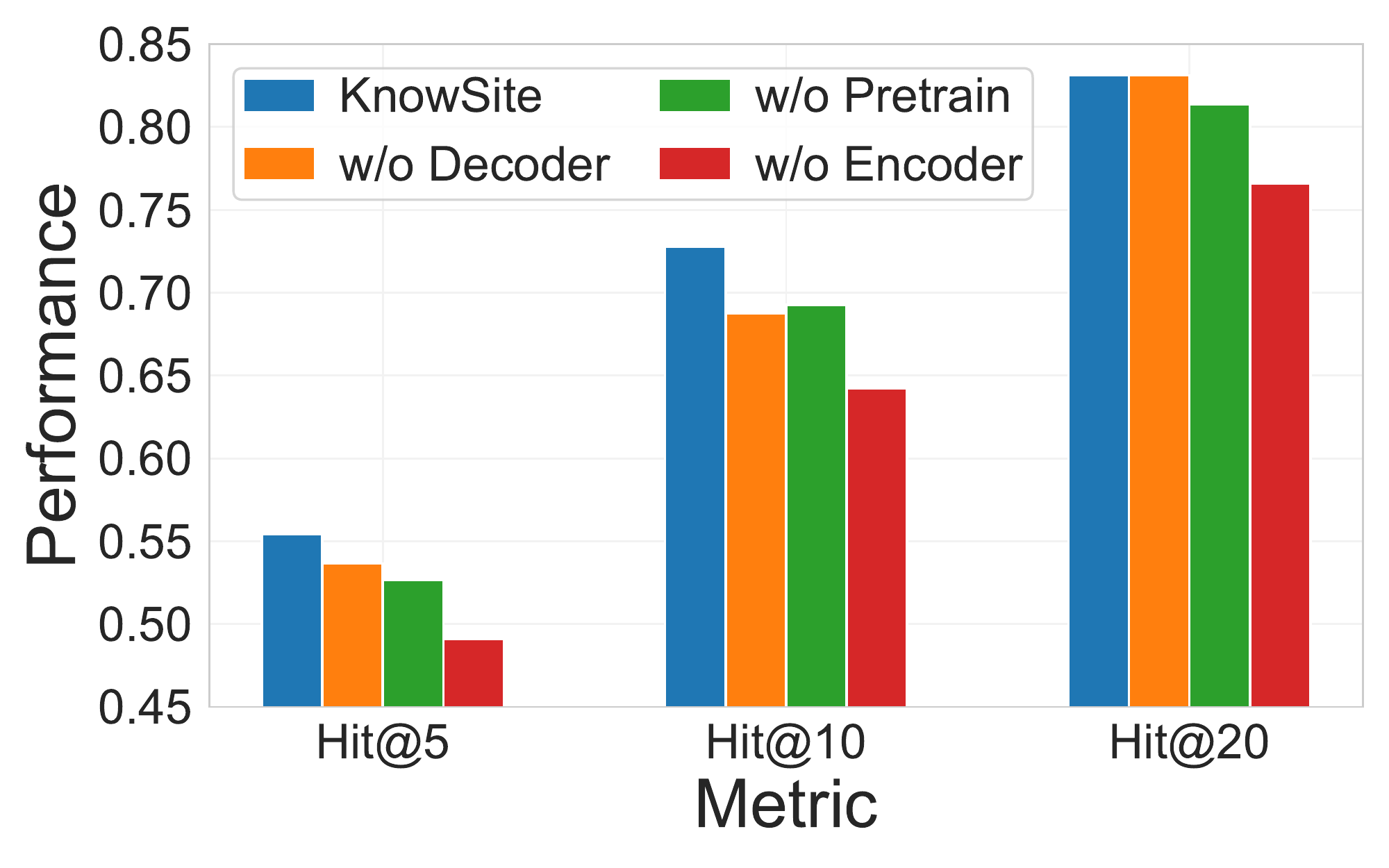}}
		\subfigure[Shanghai]{
			\includegraphics[width=0.231\textwidth]{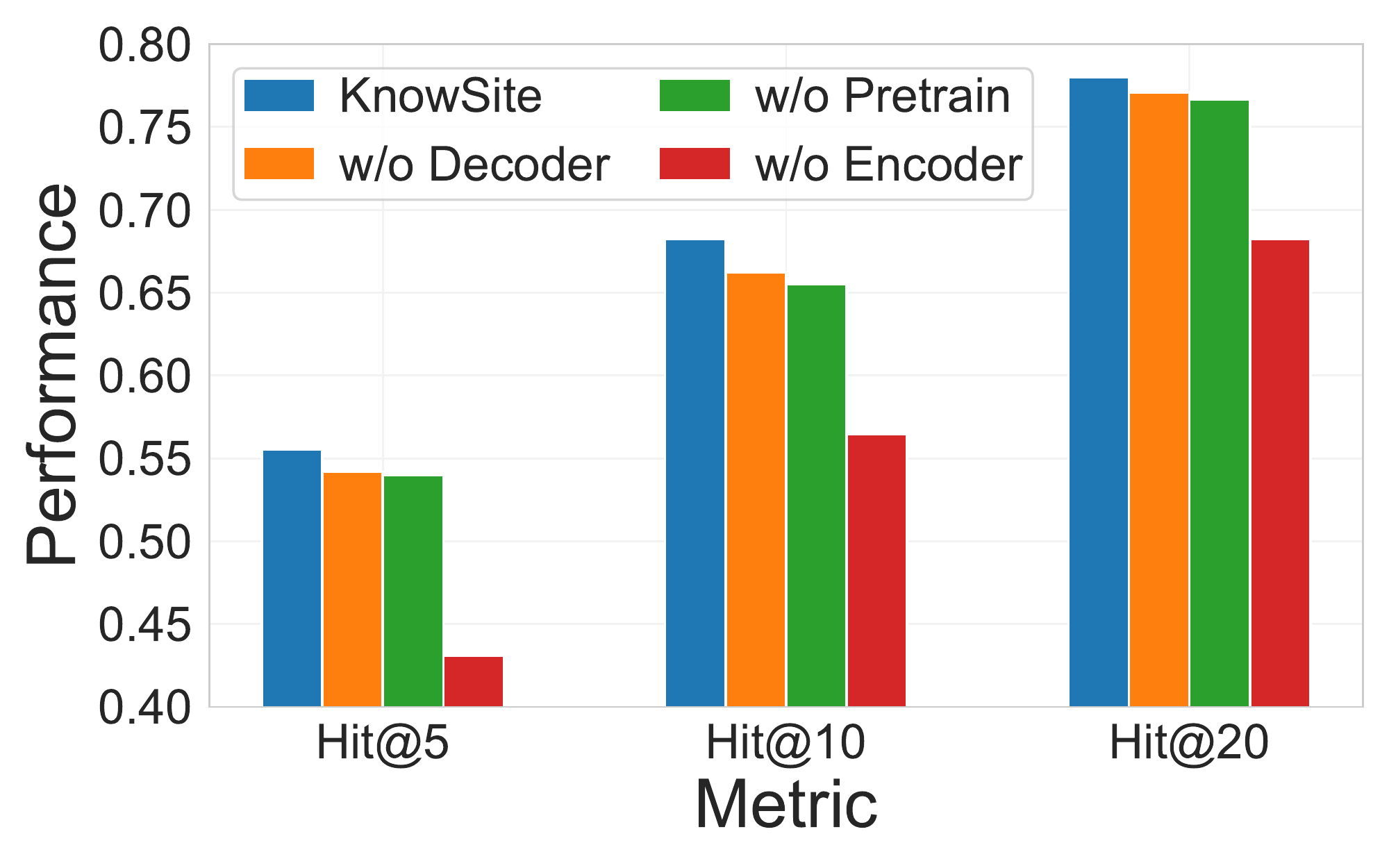}}
		\vspace{-12px}
		\caption{Performance comparison of different model variants on datasets.}\label{fig:ablation}
		\vspace{-5px}
	\end{figure}

	To evaluate the effectiveness of each module in KnowSite, 
	Figure~\ref{fig:ablation} shows the hit ratio performance of different model variants on both datasets. 
	Specifically, 
	we evaluate the KnowSite model without pre-training, GNN based encoder and relation path based decoder, respectively. 
	Note that the variant without decoder (w/o Decoder) is equivalent to the KnowSite model with $\alpha=1$ in \eqref{eq:score}.
	
	According to the results, 
	without the GNN based encoder, 
	the model performance is reduced by 12\% and 17\% on Hit@10 for Beijing and Shanghai datasets, respectively. 
	Thus, the GNN based encoder plays a quite important role in performance guarantee, 
	which confirms the importance of knowledge refinement and the gain of task-specific message passing mechanism. 
	Compared with other KRL methods, 
	the GNN based encoder successfully models diverse knowledge with site selection, 
	making the KnowSite model expressive. 
	Besides, 
	the pre-training step provides a task-agnostic but semantic initialization, 
	contributing a performance gain of 5\% on Hit@10 for datasets. 
	Moreover, relation path based decoder further achieves 4\%-5\% improvement on Hit@10 with brand-specific choice of site selection criteria. 
	Therefore, all three modules of pre-training, GNN based encoder and relation path based decoder are quite essential for effective site decisions.
	

	\subsection{Explainability Study}
	To further investigate the influence of relation paths in KnowSite as well as understand the reasons behind different brands' site decisions, 
	we present several case studies in this part. 
	
	\subsubsection{Influence of Relation Paths}
\begin{figure}[hbtp]
		\vspace{-10px}
		\centering
		\subfigure[Beijing]{
			\includegraphics[width=0.231\textwidth]{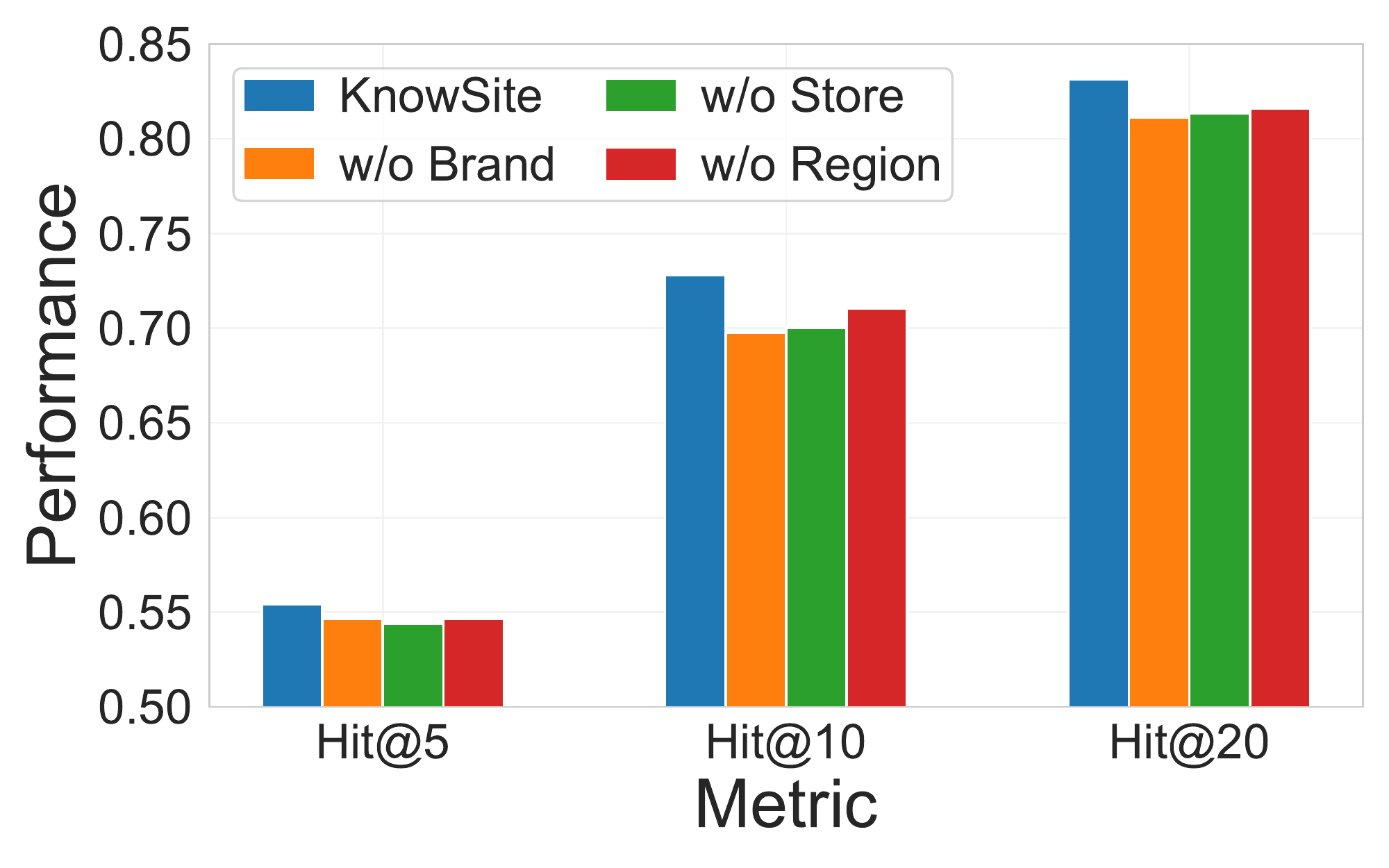}}
		\subfigure[Shanghai]{
			\includegraphics[width=0.231\textwidth]{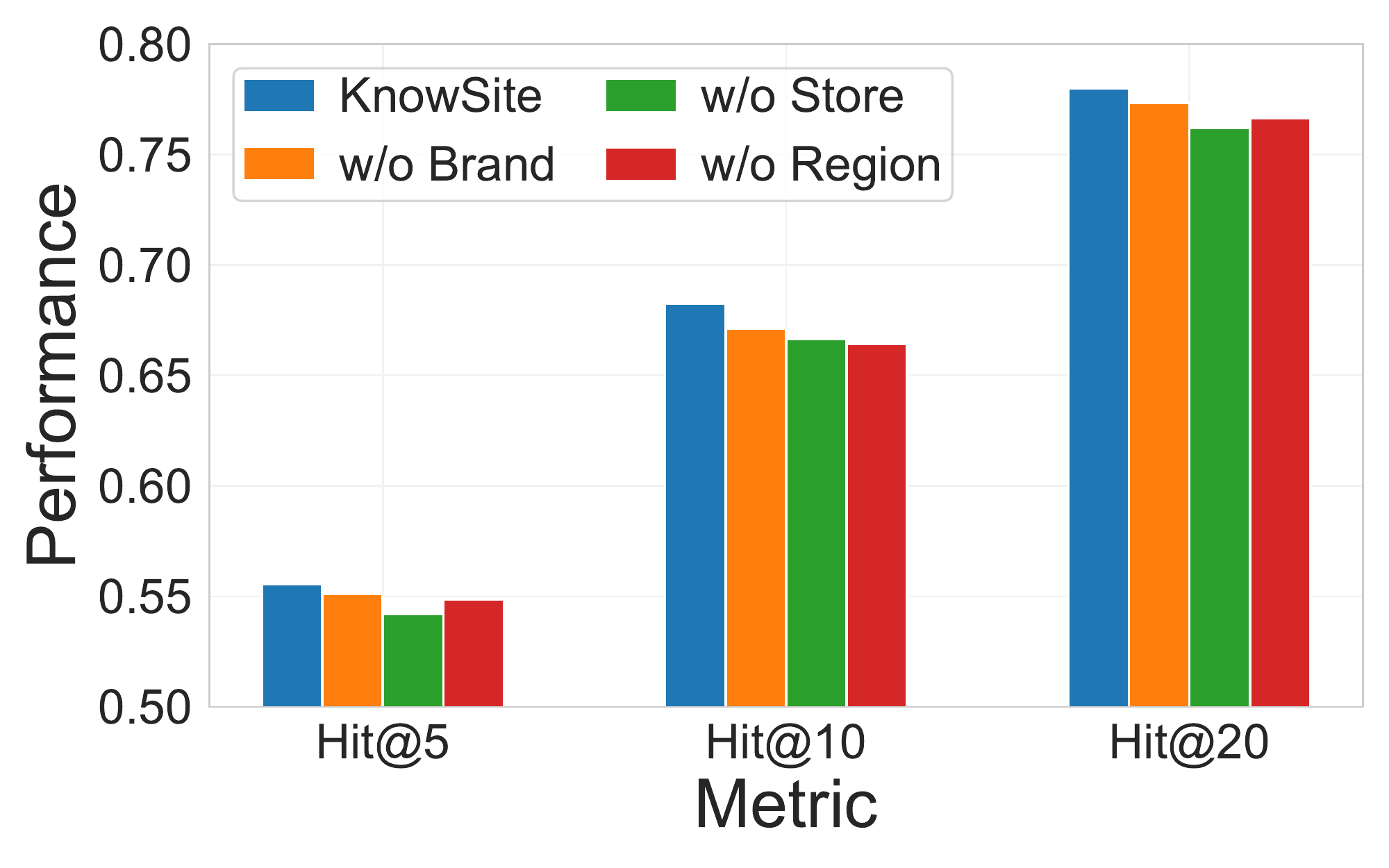}}
		\vspace{-12px}
		\caption{Performance comparison of KnowSite models without different types of relation paths on datasets.}\label{fig:rm_path}
		\vspace{-10px}
	\end{figure}

	The relation paths in Table~\ref{tab:relation_path} can be categorized into three types of region-based (the first four paths), brand-based (the 5th and 6th paths), and store-based (the last two paths) criteria, 
	and we investigate their influence on model performance by removing any type of relation paths in decoder of KnowSite, as shown in Figure~\ref{fig:rm_path}. 
	
	Overall, we observe the performance decrease in both datasets. 
	For example, 
	based on the evaluation metric of Hit@10, 
	removing brand-based relation paths brings a drop of 4\% for Beijing dataset, 
	while removing region-based ones brings a drop of 3\% for Shanghai dataset. 
	More importantly, based on results in two datasets, we are able to identify different preferences to above relation paths for different cities, which may be caused by different city structures and other social factors. 
	Specifically, 
	the region-based relation paths are the most important type for Shanghai but the least important one for Beijing. 
	This may partly owe to the different region structures. 
	Due to numerous waterways in Shanghai, 
	the regions are in irregular structure and thus own various functions and sizes, 
	which further becomes a quite important factor for site selection. 
	In contrast, 
	regions in Beijing are arranged in grid structure with similar functions and sizes, 
	which is less important than other factors like the characteristics of brands and stores. 
	Hence, the influence of relation paths provides explainable site decisions in different cities.

	\subsubsection{Brands v.s. Site Selection Criteria}
\begin{figure}[hbtp]
		\vspace{-10px}
		\centering
		\subfigure[Beijing]{
			\includegraphics[width=0.47\textwidth]{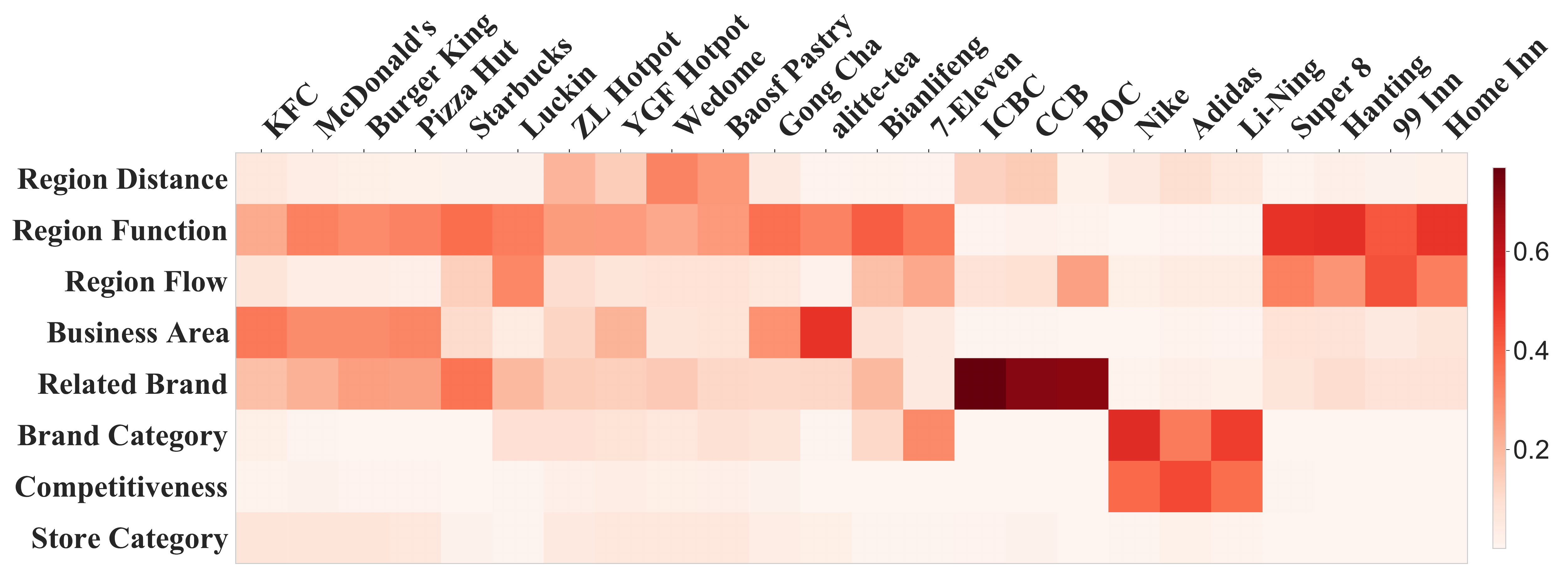}}
		\vfill
		\vspace{-1em}
		\subfigure[Shanghai]{
			\includegraphics[width=0.47\textwidth]{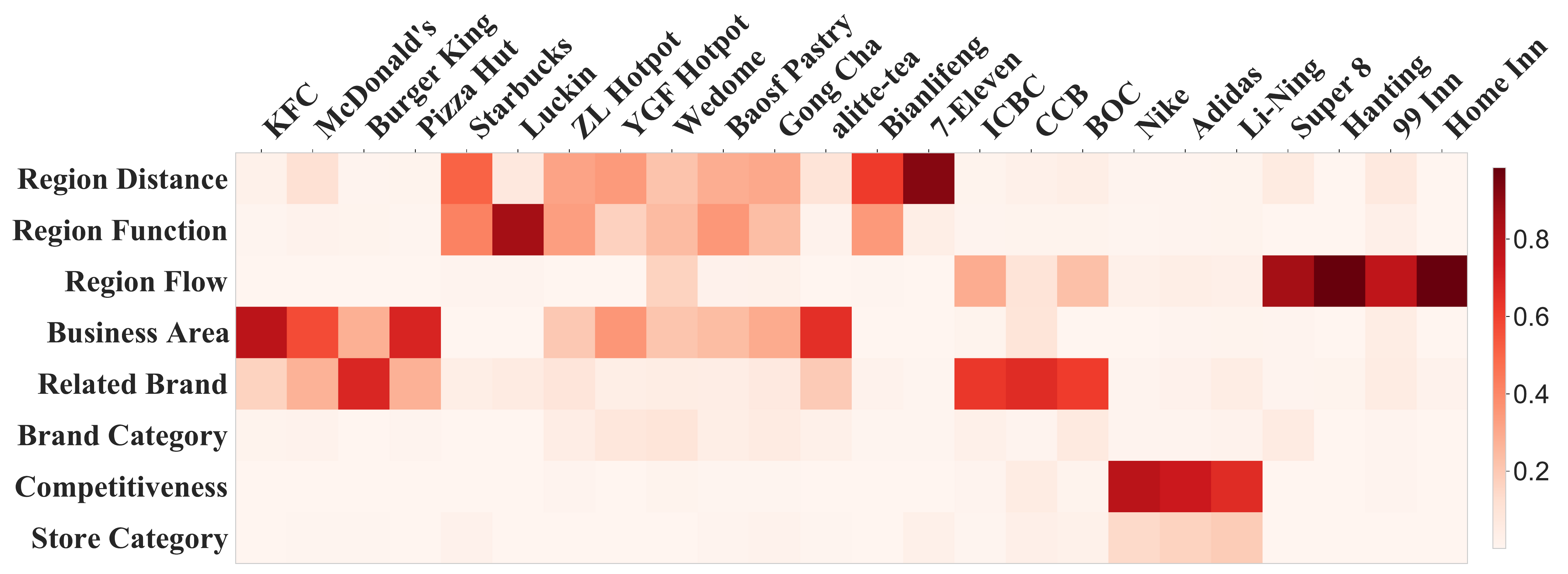}}
		\vspace{-12px}
		\caption{Attention weight visualization of different brands to site selection criteria on datasets.}\label{fig:attn_brand}
		\vspace{-10px}
	\end{figure}

	As described in Section~\ref{sec:decoder}, 
	the attention weights in \eqref{eq:attn} show the relationship between brands and criteria. 
	Thus, we present the attention weight visualization on two datasets in Figure~\ref{fig:attn_brand}. 
	Several typical brands across food, leisure sports, accommodation and other categories are selected for visualization. 
	A description of selected brands can be found in Section~\ref{app:select_brands} for better understanding.
	
	By combining visualization results in different cities, i.e., Figure~\ref{fig:attn_brand}(a) and (b) together, we have similar findings regarding to brands' preference to selection criteria that are both insightful and convincing. 
	First, all fast-food chain brands like KFC, McDonald's, Burger King and Pizza Hut determine optimal locations with business area condition and related brand strategy considered, 
	which is in accord with the location game between brands \cite{talwalkar_2012} as well as the commonsense that there always is one KFC store near one McDonald's store \cite{hu2013comparative}. 
	Second, similar attention on related brand strategy can also be observed among bank brands of ICBC, CCB and BOC, three large banks in China. 
	These bank brands also focus on region flow for more customers. 
	Moreover, the last four columns in figures represent the preference of four popular hotel chain brands to region flow, which determines the occupancy directly. 
	Note that the slight difference between results in Figure~\ref{fig:attn_brand}(a) and (b) may be caused by different city conditions and noise in model learning. 
	Overall, such results demonstrate the explainable capability of our proposed KnowSite model, 
	which can provide a good reference for site selection understanding.

	%
	
\begin{figure}[hbtp]
		\vspace{-10px}
		\centering
		\subfigure[Task-agnostic output]{
			\includegraphics[width=0.231\textwidth]{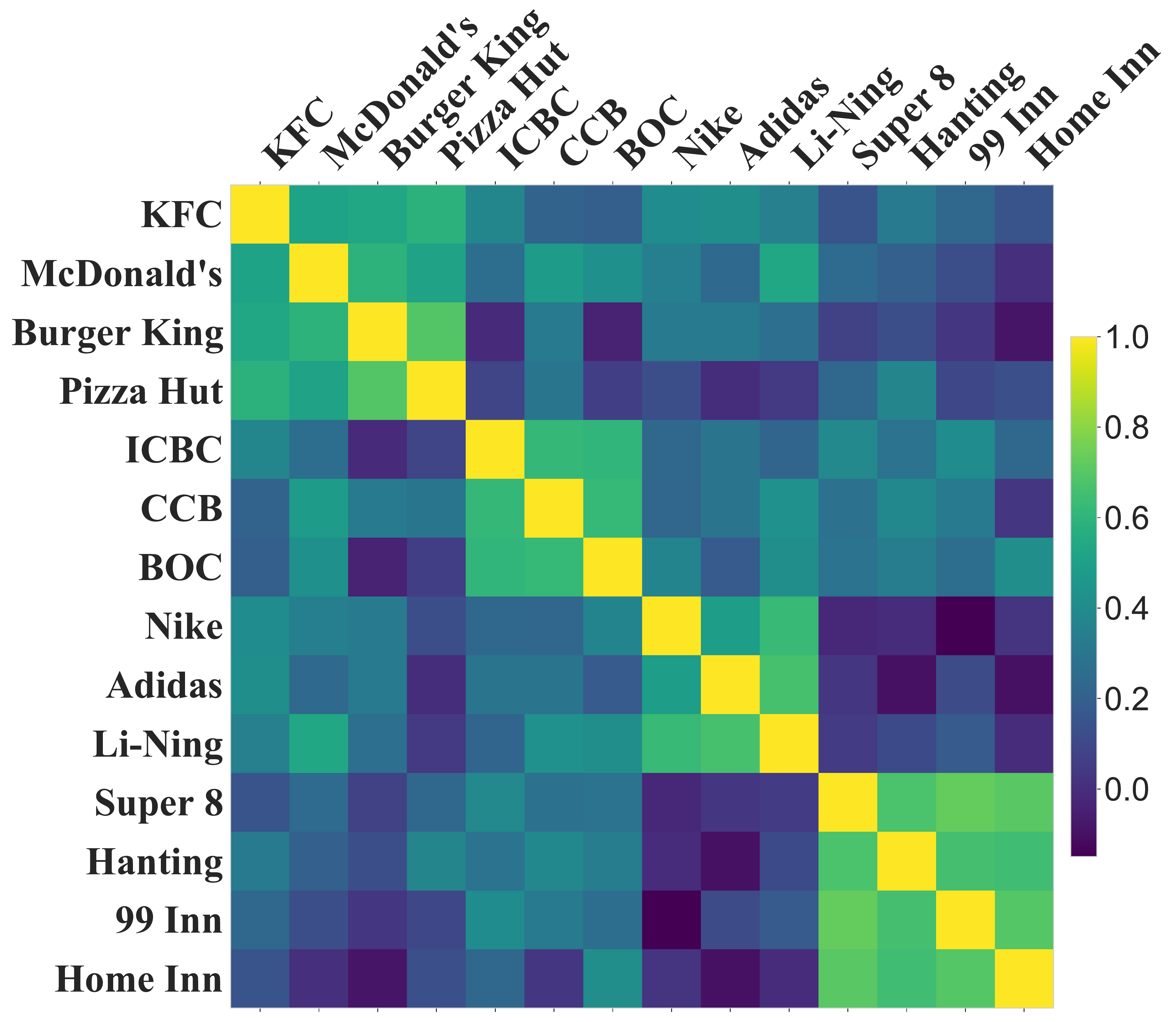}}
		\subfigure[Task-specific output]{
			\includegraphics[width=0.231\textwidth]{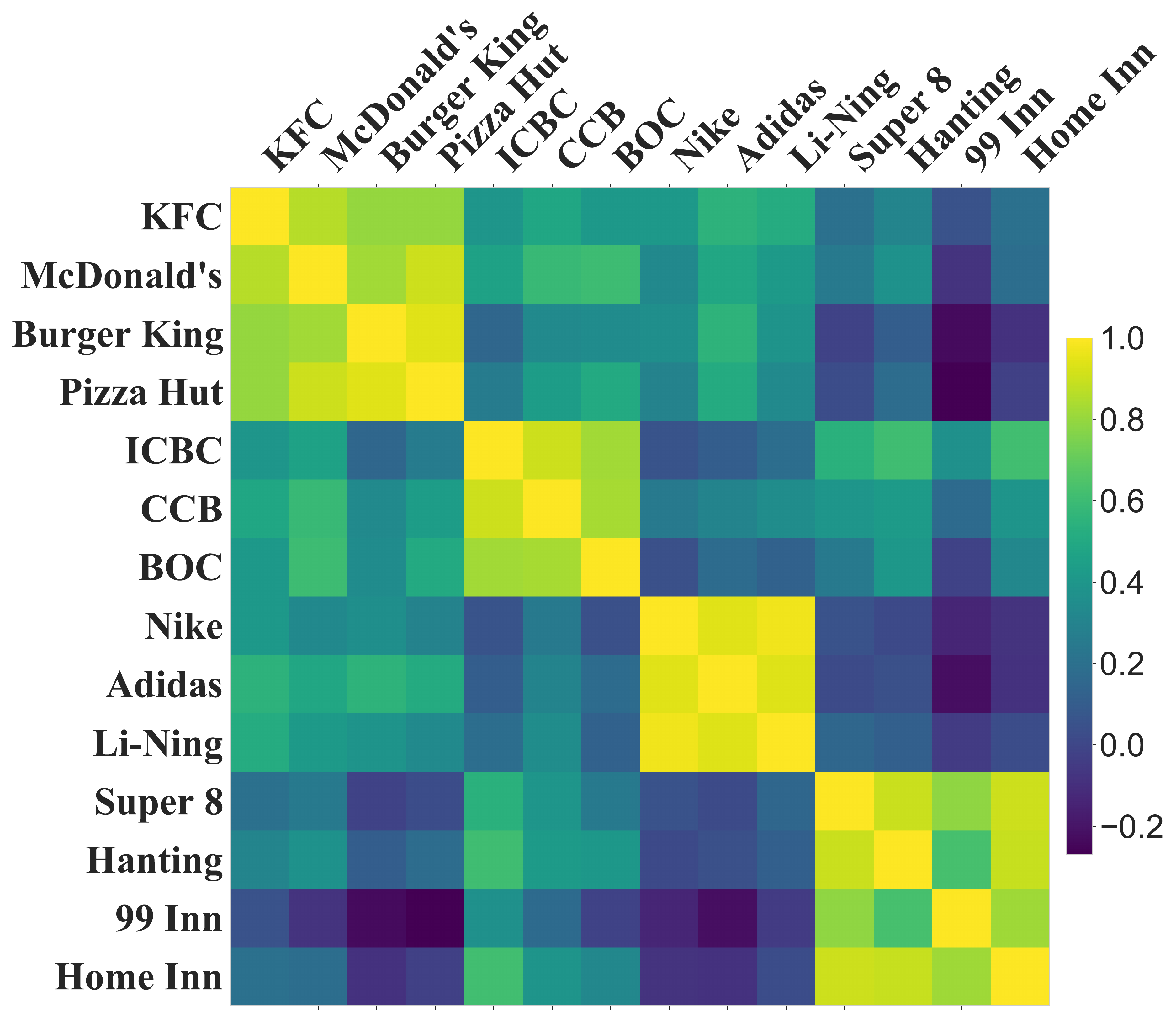}}
		\vspace{-12px}
		\caption{Cosine distance visualization of different brands' representations in Beijing.}\label{fig:brand_cosine}
		\vspace{-10px}
	\end{figure}

	To further investigate the influence of site selection criteria on
	brand representations, Figure~\ref{fig:brand_cosine} visualizes the cosine distance between selected brands in Beijing, 
	in which Figure~\ref{fig:brand_cosine}(a) utilizes task-agnostic representations  $\bm{h}^0_b$ of pre-training, 
	while Figure~\ref{fig:brand_cosine}(b) utilizes task-specific ones $\bm{h}^K_b$ of GNN based encoder output with end-to-end training. 
	Since UrbanKG contains semantic information like \emph{RelatedBrand} links, 
	related brands' representations are closer compared with others, 
	as shown in diagonal blocks of Figure~\ref{fig:brand_cosine}(a). 
	However, 
	due to the task-agnostic learning' in pre-training step, 
	such correlation is not that obvious. 
	In comparison, 
	a remarkable brand correlation is illustrated in Figure~\ref{fig:brand_cosine}(b).
	Several highlight diagonal blocks indicate the closeness of brands in hidden space, 
	such as the first block of four fast-food chain brands and the last block of four hotel chain brands. 
	Besides, 
	the brand correlations in off-diagonal parts are also enhanced in Figure~\ref{fig:brand_cosine}(b), 
	which also suggests the effectiveness of knowledge refinement with brand information encoding. 
	Therefore, 
	KnowSite successfully captures the semantic relatedness among brands and reveals the relationship between brands and various site selection criteria.

	\subsubsection{Categories v.s. Site Selection Criteria}
\begin{figure}[hbtp]
		\vspace{-10px}
		\centering
		\subfigure[Beijing]{
			\includegraphics[width=0.231\textwidth]{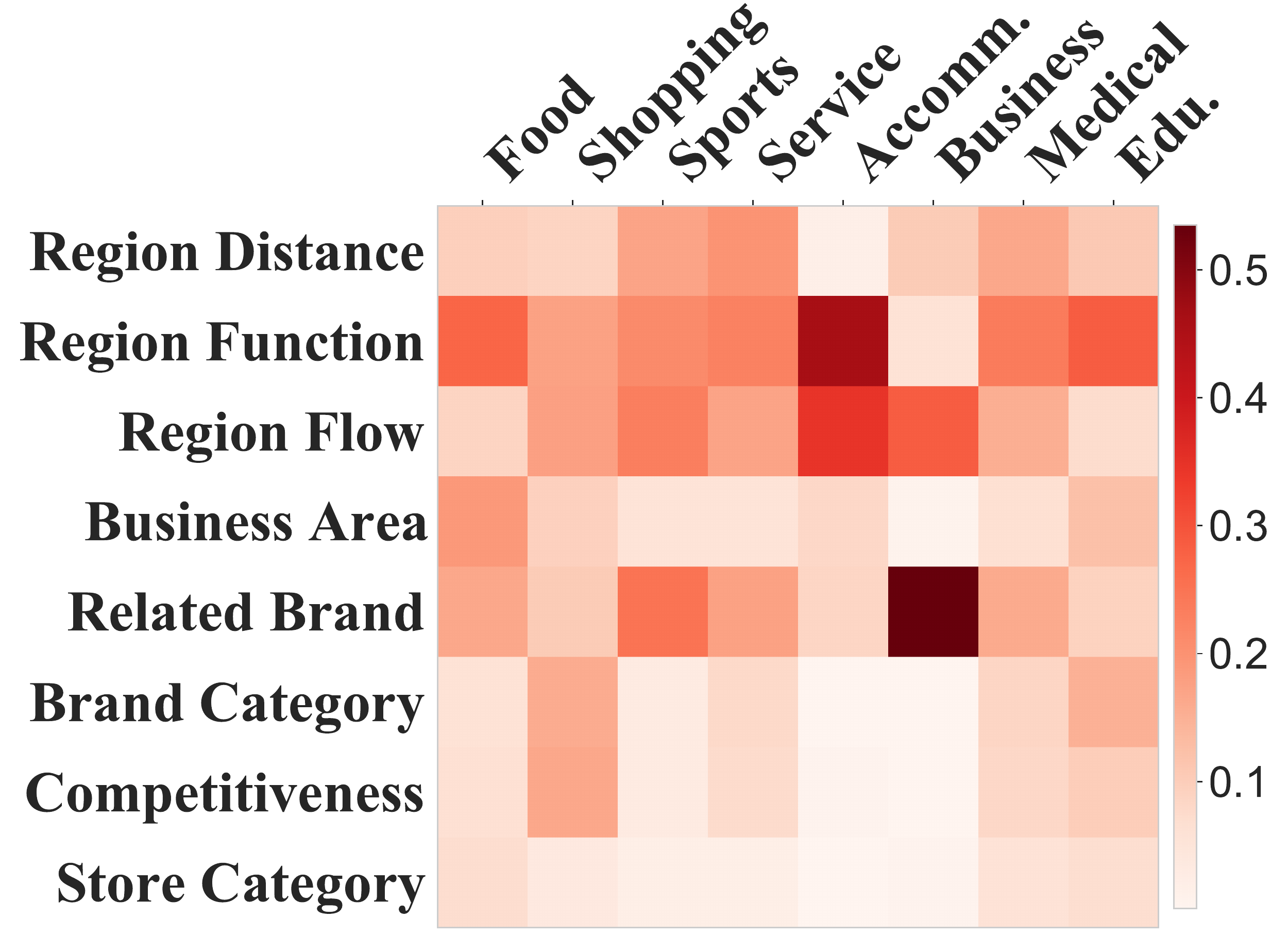}}
		\subfigure[Shanghai]{
			\includegraphics[width=0.231\textwidth]{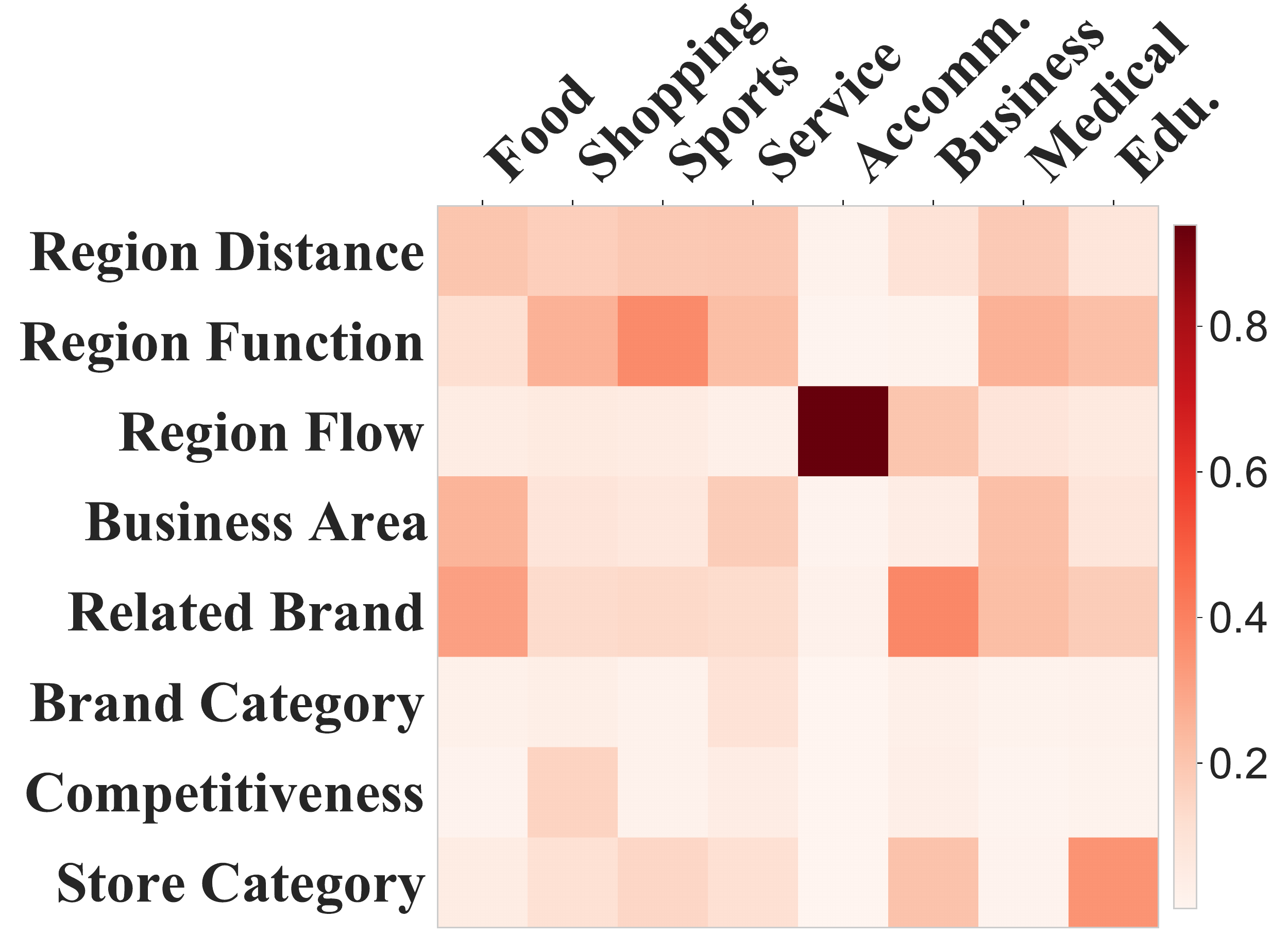}}
		\vspace{-12px}
		\caption{Attention weight visualization of different categories to site selection criteria on datasets. Sports, Service, Accomm., Edu. represent leisure sports, life service, accommodation and education, respectively.}\label{fig:attn_cate}
		\vspace{-10px}
	\end{figure}

	In Figure~\ref{fig:attn_cate}, 
	we further reveal the relationship between categories and site selection criteria. 
	For each dataset, 
	eight typical categories are selected, 
	and the attention weights of all brands under corresponding categories are averaged for visualization. 
	
	Similar phenomenons in Figure~\ref{fig:attn_brand} can be observed in Figure~\ref{fig:attn_cate}. 
	The brands of food category focus on business area and related brand, 
	while the brands of accommodation category pay more attention to region function and flow. 
	Among the site selection criteria, 
	the region factors of distance and function as well as related brand strategy are commonly considered across various categories. 
	Such results again validate the importance of relation path based decoder in KnowSite, 
	and explore its potential in site selection for both brand and category levels. 
	
	Throughout the experimental study, 
	KnowSite achieves the state-of-the-art performance on site selection task, 
	and the effectiveness of each designed module is validated. 
	Moreover, 
	with relation paths and attention mechanism utilized, 
	KnowSite successfully reveals the influences of site selection criteria on various businesses. 
	

	\section{Related Work}\label{sec:related_work}
	Here we discuss some closely related studies, 
	including site selection methods, KRL with KG and KG applications in urban computing. 
	
	With multi-source urban data available, 
	the data-driven methods first extract features from data, 
	and then learn regression/learning-to-rank models for the problem \cite{geo-spotting, xu2016demand}. 
	Specifically, 
	both Geo-Spotting \cite{geo-spotting} and DD3S \cite{xu2016demand} firstly investigate the predictive power of various features like density, competitiveness and area popularity, 
	and then apply traditional SVR \cite{svr} and RankNet \cite{ranknet} to determine the optimal location. 
	However, 
	these methods learn individual models for each brand and cannot generalize to various businesses. 
	Furthermore, several works also integrate deep network with feature engineering \cite{liu2019deepstore, xu2020ar2net,li2018commercial}. 
	For example, 
	DeepStore \cite{liu2019deepstore} and AR$^2$Net \cite{xu2020ar2net} extracts features from commercial data, satellite images, etc., and further combine deep neural networks with attention mechanism for solution. 
	UKG-NN \cite{zhang2018structured} builds a relational graph with manually defined features, 
	which are passed to the neural network for site decisions. 
	NeuMF-RS \cite{li2018commercial} adds restaurants' and sites' attributes to neural collaborative filtering for site selection. 
	However, deep models suffer from explainability issues with black box neural networks. 
	Both traditional and deep models fail to extract vital information from data according to site selection criteria, i.e., knowledge, 
	whose performance is easily affected by the quality of upstream feature engineering. 
	In contrast, our work leverages knowledge-driven paradigm for both effective and explainable performance. 
	
	
	As for KRL to learn embeddings of entities and relations, though complete structures like GNN have been introduced \cite{vashishth2019composition,schlichtkrull2018modeling}, 
	tensor decomposition models still achieve the best performance \cite{ji2021survey}, such as DistMult \cite{yang2014embedding}, ComplEx \cite{trouillon2016complex} and TuckER \cite{balazevic2019tucker}. 
	Here we argue that the proposed GNN encoder is more suitable for representing specific knowledge of site selection, as it can flexibly control the information sharing among diverse factors. 
	Meanwhile, multi-hop relation paths have been introduced in KRL for more accurate representations \cite{lin2015modeling,zhu2019representation}. 
	In the proposed KnowSite, we adopt relation path based decoder to model site selection criteria for brands. 
	Thus, it not only boosts the performance, but also provides explainable site decisions based on the relation path logic. 
	Note that our relation path based on KG is different from the meta-path counterpart in heterogeneous graphs \cite{yang2020heterogeneous}, which only learns node embeddings but ignores edge representations \cite{wang2019heterogeneous}, thus not applicable to this work.

	In addition, there are some attempts to apply KG for urban computing. 
	For example, 
	the construction of geographic KGs is investigated in \cite{sun2019demonstrating,yan2019spatially}, 
	where the spatial relationships between geographic components are extracted. 
	Some works \cite{wang2020incremental,dadoun2019location,zhang2016semantic} introduce KG with two or three relations and ontologies for specific applications. 
	However, such developed KGs miss important knowledge for site selection such as human flow, competitiveness, brand relatedness, etc. 
	In comparison, 
	our proposed UrbanKG contains rich site selection related knowledge with over 20k entities in the city and over 300k facts between them, which is a promising backbone for various applications in urban computing.

	\section{Conclusion}\label{sec:conclusion}
	In this work, 
	we proposed KnowSite, 
	a knowledge-driven model for site selection. 
	By leveraging KG for urban knowledge representation, 
	KnowSite develops a generalized encoder-decoder framework, 
	where multi-relational message passing and criteria-based relation paths are adopted to reason different brands' site decisions. 
	Extensive experiments demonstrate that KnowSite achieves superior performance with both effectiveness and explainability achieved.
	
	As future work, 
	we will combine KnowSite with the traditional data-driven paradigm, 
	and utilize both KRL methods and feature engineering towards powerful site selection. 
	Moreover, 
	we plan to explore our proposed UrbanKG as well as the generalized encoder-decoder framework for other urban computing tasks such as flow prediction, mobility prediction, etc.
	
	\newpage
	\bibliographystyle{ACM-Reference-Format}
	\balance
	\bibliography{sample-base}
	\newpage
	\appendix
	
	\section{Details of Dataset}\label{app:data_details}
	\subsection{Dataset Statistics}
	Two datasets are built for evaluation:
	\begin{itemize}
		\item \textbf{Beijing}: This dataset focuses on the area within the Fifth Ring Road, Beijing, China. 
		\item \textbf{Shanghai}: This dataset focuses on the whole area of Shanghai, China.
	\end{itemize}
	The brands with over 20 stores are selected for dataset construction, 
	and the site selection data are randomly split into train/valid/test sets by a proportion of 6:2:2. 
	For brands opening multiple stores at one region, we aggregate corresponding store data into one brand-region sample. 
	Moreover, the UrbanKGs are developed following the construction in Section~\ref{sec:urbankg_construction}, 
	The number of clicks in $\mathcal{D}_\textnormal{Click}$ is utilized to indicate the popularity at regions. 
	The statistics of the dataset and the UrbanKG are summarized in Table~\ref{tab:dataset}. 
	\begin{table}[htbp]
		\caption{Dataset statistics. \#triplet denotes the number of triplets in the corresponding UrbanKGs.}\label{tab:dataset}
		\vspace{-10px}
		\setlength\tabcolsep{1.5pt}
		\begin{tabular}{ccccccccc}
			\toprule
			Dataset     & $\left|\mathcal{E}\right|$    & $\left|\mathcal{R}\right|$ & \#triplet & \#Brand & \#Region & \#Train & \#Valid  & \#Test \\
			\midrule
			Beijing  & 23,754 & 35  & 330,652    & 398     & 528      & 15,022  & 5,007 & 5,008  \\
			Shanghai & 41,338 & 36  & 589,852    & 441     & 2,042     & 29,006  & 9,669 & 9,669 	\\
			\bottomrule     
		\end{tabular}
		\vspace{-10px}
	\end{table}
	
	Table~\ref{tab:ont} introduces the ontology statistics of UrbanKG, 
	i.e., 
	the number of entities the for corresponding ontology. 
	As for POIs in the construction of UrbanKG, we only consider those belonging to selected brands in datasets. 
	\begin{table}[H]
		\caption{The ontology statistics of UrbanKG for cities.}\label{tab:ont}
		\vspace{-10px}
		\setlength\tabcolsep{2pt}
		\begin{tabular}{ccccccccccc}
			\toprule
			Dataset     & \#Brand & \#Region & \#Ba & \#POI  & \#1-Cate & \#2-Cate & \#3-Cate \\
			\midrule
			Beijing   & 398     & 528      & 168  & 22,468 & 10       & 39       & 143      \\
			Shanghai  & 441     & 2042     & 264  & 38,394 & 11       & 42       & 144	\\
			\bottomrule     
		\end{tabular}
	\end{table}
	
	Table~\ref{tab:relations_number} shows the relational fact statistics of UrbanKG in two cities for our work.
	
	\begin{table}[htbp]
		\caption{The details of defined relations in UrbanKG.}\label{tab:relations_number}
		\vspace{-10px}
		\begin{tabular}{c|c|c}
			\toprule
			Relation &  Beijing &  Shanghai \\
			\midrule
			\emph{BorderBy} &  2,626 &  9,896 \\
			\emph{NearBy} & 7,232  & 29,942  \\
			\emph{FlowTransition} & 287 & 634\\
			\emph{SimilarFunction} &  2,844  &  5,126 \\
			\emph{Competitive} & 1,968  &   2,576  \\
			\emph{RelatedBrand} & 296 &  352 \\
			\emph{SubCateOf\_ij} & 325 &  330 \\ 
			\emph{BaServe} &  6,152  & 11,876 \\
			\emph{BelongTo} & 22,372  &  38,394   \\
			\emph{LocateAt}  & 22,468 &  38,394     \\
			\emph{POIToCate\_i} & 22,468*3  &  38,394*3 \\
			\emph{BrandToCate\_i} & 398*3  &  441*3 \\
			\emph{BrandOf} & 22,468  & 38,394 \\
			\emph{OpenStoreAt} &  15,022  & 29,006  \\
			\bottomrule
		\end{tabular}  
	\end{table}

	\subsection{Details of Selected Brands for Visualization}\label{app:select_brands}
	Here we give a description of selected brands in experiments. 
	\begin{itemize}
		\item \textbf{KFC}, \textbf{McDonald's}, \textbf{Burger King}, \textbf{Pizza Hut}. 
		Fast-food chain brands around the world. 
		\item \textbf{Starbucks}, \textbf{Luckin}. Coffeehouse chain brands. Luckin, founded in Beijing, manages more stores than Starbucks in China.
		\item \textbf{ZL(Zhangliang) Spicy Hotpot}, \textbf{YGF(Yang Guofu) Spicy Hotpot}. Two of the largest spicy hotpot (a.k.a. Mala Tang, Chinese snack) chain brands in China.
		\item \textbf{Wedome}, \textbf{Baosf Pastry}. Bakery chain brands in China, focus on cakes, bread, and bakery items.
		\item \textbf{Gong Cha}, \textbf{alittle-tea}. Tea chain brands, offering both original tea and milk tea.
		\item \textbf{Bianlifeng}, \textbf{7-Eleven}. Convenience store chain brands. 	
		\item \textbf{ICBC (Industrial and Commercial Bank of China)}, \textbf{CCB (China Construction Bank)}, \textbf{BOC (Bank of China)}. State-owned commercial bank companies in China, opening branch banks and ATMs throughout the country.
		\item \textbf{Nike}, \textbf{Adidas}, \textbf{Li-Ning}. Leisure sport chain brands. 
		\item \textbf{Super 8 (Hotel)}, \textbf{Hanting (Hotel)}, \textbf{99 Inn}, \textbf{Home Inn}. Four of the largest hotel chain brands in China.
	\end{itemize}
	
	\begin{figure*}[hbtp]
		\centering
		\subfigure[Pre-training output]{
			\includegraphics[width=0.331\textwidth]{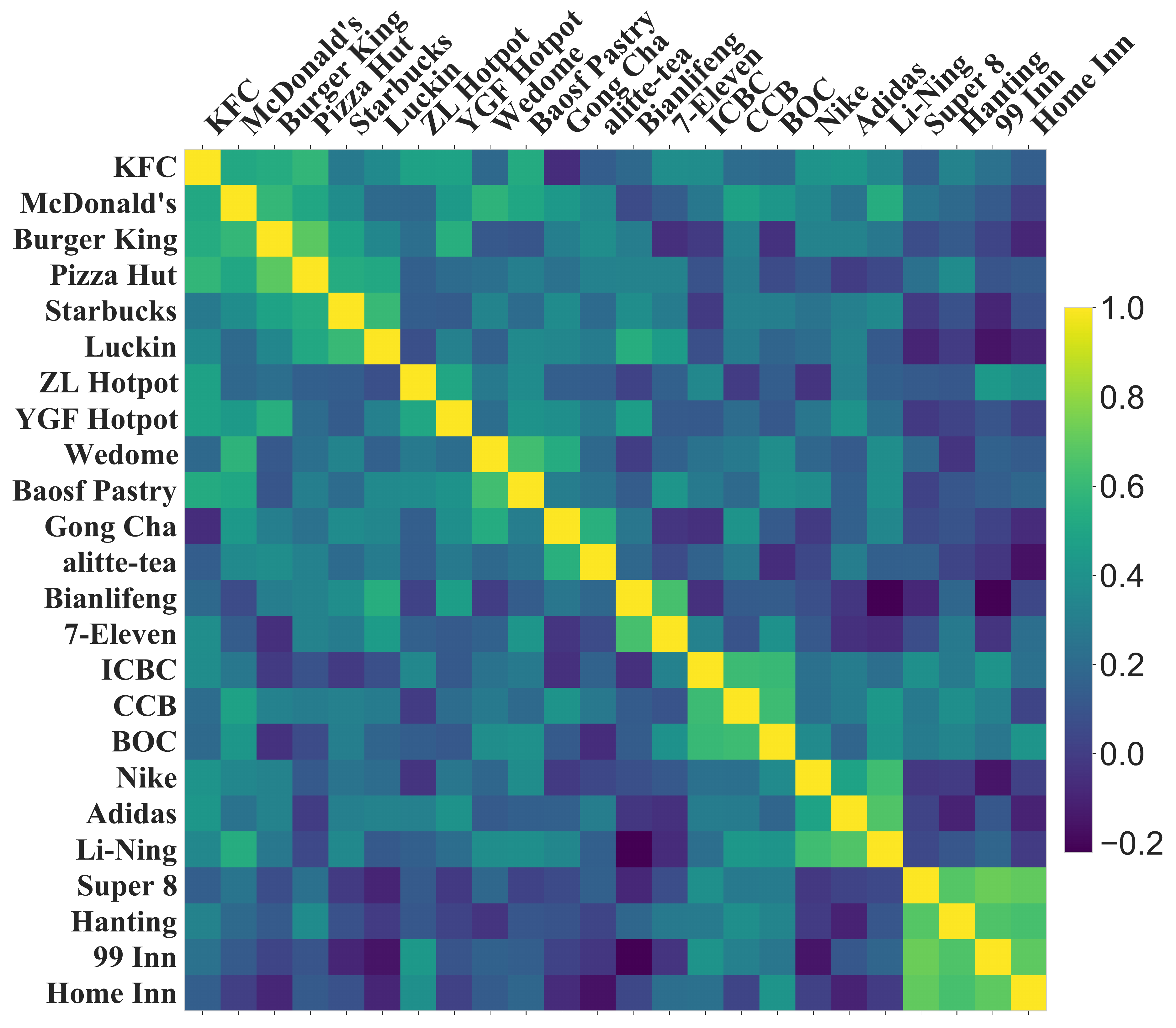}}
		\subfigure[Encoder output]{
			\includegraphics[width=0.331\textwidth]{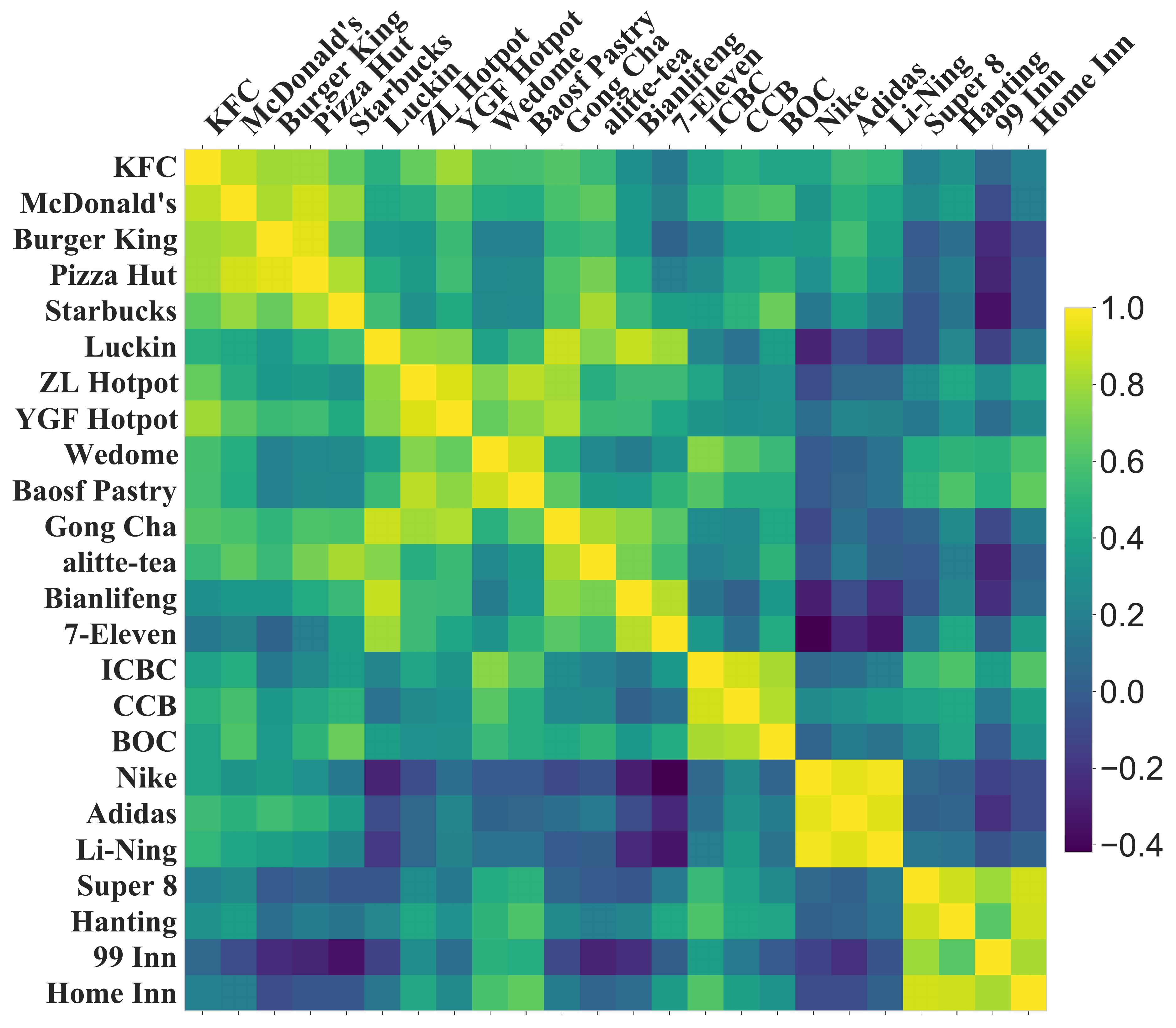}}
		\vspace{-10px}
		\caption{Cosine distance visualization of different brands' representations in Beijing.}\label{fig:brand_cosine_beijing_all}
	\end{figure*}
	
	\begin{figure*}[hbtp]
		\centering
		\subfigure[Pre-training output]{
			\includegraphics[width=0.331\textwidth]{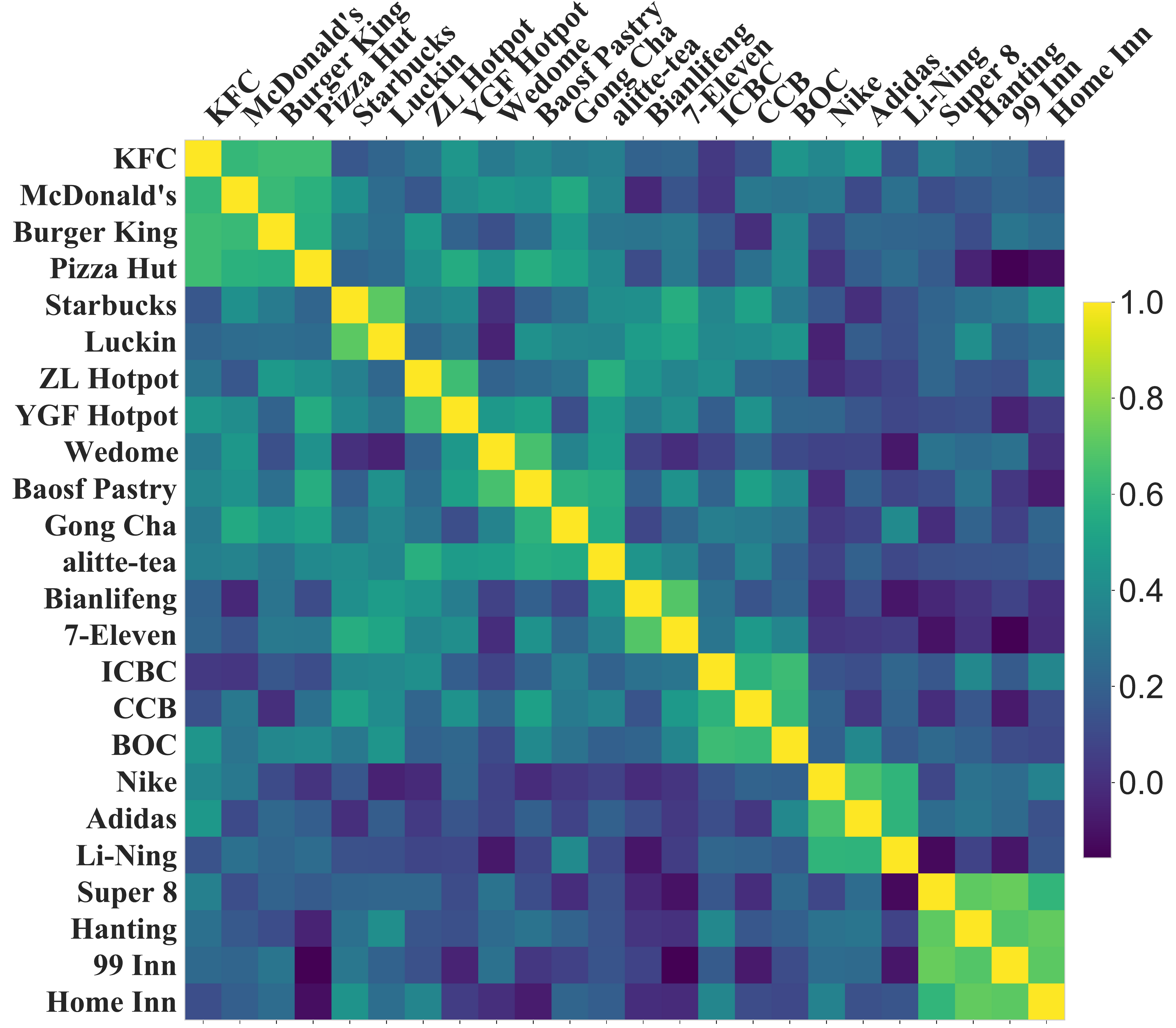}}
		\subfigure[Encoder output]{
			\includegraphics[width=0.331\textwidth]{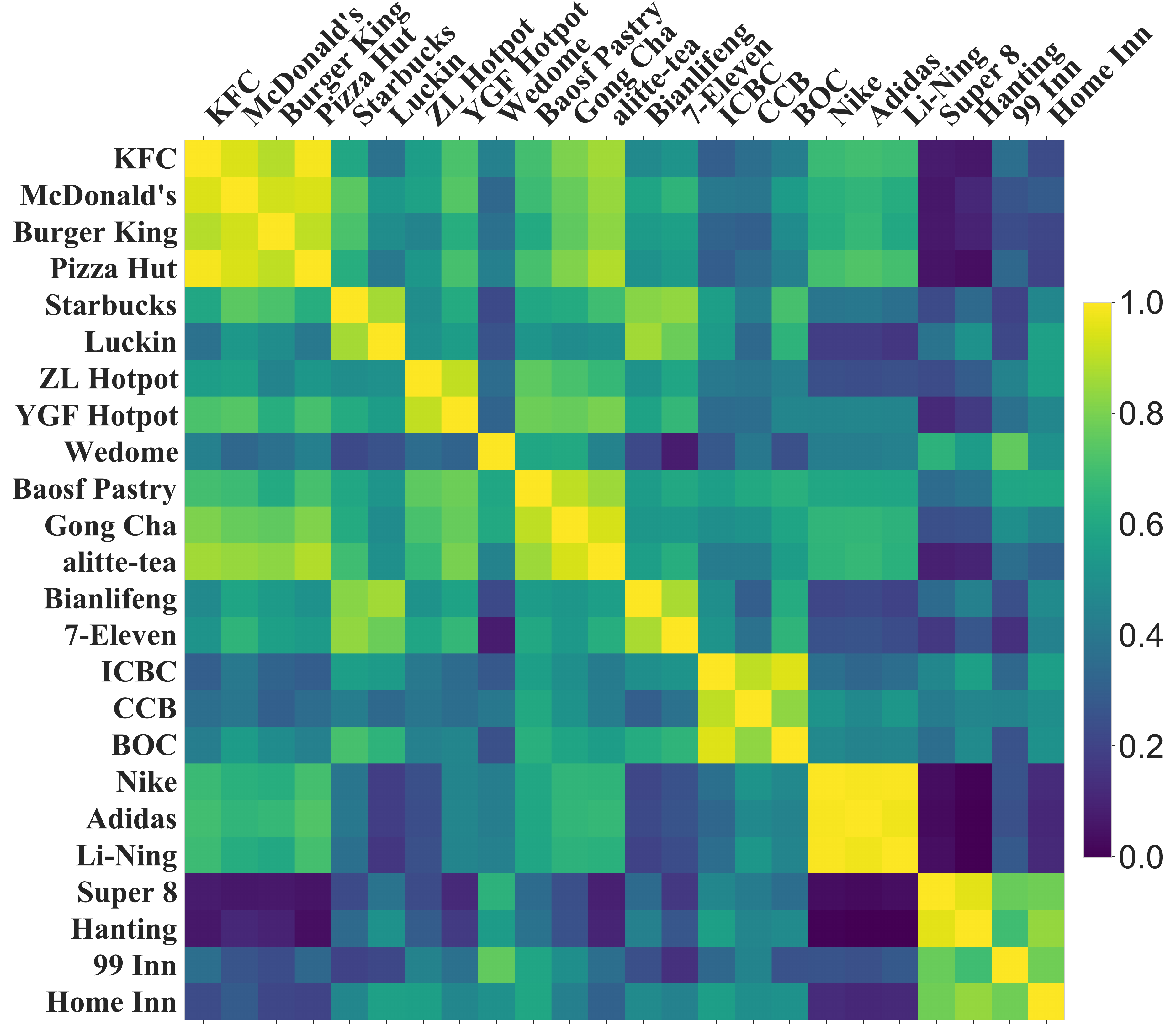}}
		\vspace{-10px}
		\caption{Cosine distance visualization of different brands' representations in Shanghai.}\label{fig:brand_cosine_shanghai_all}
	\end{figure*}

	\section{Experimental Details} \label{app:exp}
	\subsection{Metrics} \label{app:metrics}
	Given the region set $\mathcal{A}$, the brand set $\mathcal{B}$ and the $i$-th brand, we denote $A^i$ and $\hat{A}^i$ as its true and model predicted region list based on popularity/predicted score, respectively. 
	$n_i$ denotes the number of regions in test set where the $i$-th brand opens the store. Then the metrics are calculated as follows,
	\begin{itemize}
		
		\item $\textnormal{NDCG}@k$ (Normalized Discounted Cumulative Gain), which measures the extent to which the top-k regions in $A^i$ are highly ranked in $\hat{A}^i$. 
		\begin{align}
			\textnormal{NDCG}@k=\frac{1}{\left|\mathcal{B}\right|}\sum^{\left|\mathcal{B}\right|}_{i=1}\textnormal{NDCG}_i@k, 
			\ \ \textnormal{DCG}_i@k=\sum^k_{j=1}\frac{2^{rel(\hat{a}_j)}-1}{\log_2(j+1)}, \notag
		\end{align}
		where the relevance score $rel(\hat{a}_j)$ follows the definition in \cite{geo-spotting}, i.e., 
		$rel(\hat{a}_j)=\frac{\left|\mathcal{A}\right|-rank(\hat{a}_j)+1}{\left|\mathcal{A}\right|}$ for ground truth and $rel(\hat{a}_j)=0$ for invalid regions.  
		$\textnormal{NDCG}_i@k$ is obtained by normalizing $\textnormal{DCG}_i@k$ via the ideal prediction $\textnormal{IDCG}_i@k$.
		
		\item $\textnormal{Hit}@k$, which describes the hit ratio of top-k regions in $A^i$. 
		\begin{align}
			\textnormal{Hit}@k=\frac{1}{\left|\mathcal{B}\right|}\sum^{\left|\mathcal{B}\right|}_{i=1}\mathbb{I}(\left|A^i_{1:k}\cap\hat{A}^i_{1:k}\right|), \notag
		\end{align}
		where $\mathbb{I}(\cdot)$ denotes the indicator function, i.e., $\mathbb{I}(x)=1$ if $x>0$, otherwise $\mathbb{I}(x)=0$.
		
		\item $\textnormal{Precision}@k$ and $\textnormal{Recall}@k$, which are defined as follows, 
		\begin{align}
			\textnormal{Precision}@k=\frac{1}{\left|\mathcal{B}\right|}\sum^{\left|\mathcal{B}\right|}_{i=1}\frac{|A^i\cap\hat{A}^i_{1:k}|}{k}. \notag \\
			\textnormal{Recall}@k=\frac{1}{\left|\mathcal{B}\right|}\sum^{\left|\mathcal{B}\right|}_{i=1}\frac{|A^i\cap\hat{A}^i_{1:k}|}{\min(n_i,k)}. \notag
		\end{align}
		
		\item $\textnormal{MAP}@k$ (Mean Average Precision), which measures the relative ranking quality of the top-k regions in $\hat{A}^i$. 
		\begin{align}
			\textnormal{MAP}@k=\frac{1}{\left|\mathcal{B}\right|}\sum^{\left|\mathcal{B}\right|}_{i=1}\frac{1}{\min(n_i,k)}\cdot\sum_{j=1}^{k}\frac{|A^i\cap\hat{A}^i_{1:j}|}{j}\cdot rel(\hat{a}_j), \notag
		\end{align}
		where $rel(\hat{a}_j)$ follows the same definition above. 
	\end{itemize}

	\subsection{Brands v.s. Site Selection Criteria}
	In Figure~\ref{fig:brand_cosine_beijing_all} and \ref{fig:brand_cosine_shanghai_all}, 
	we provide the complete results of cosine distance visualization of all selected brands' representations in Beijing and Shanghai, respectively. 
	According to the results, the relatedness between brands are further refined by GNN based encoder in KnowSite, i.e., a strong correlation can be observed on Figure~\ref{fig:brand_cosine_beijing_all}(b) and \ref{fig:brand_cosine_shanghai_all}(b), 
	which further validates the effectiveness of task-specific representation learning for site selection.

	\subsection{UrbanKG Embedding Visualization}

\begin{figure}[H]
		\centering
		\includegraphics[width=.95\linewidth]{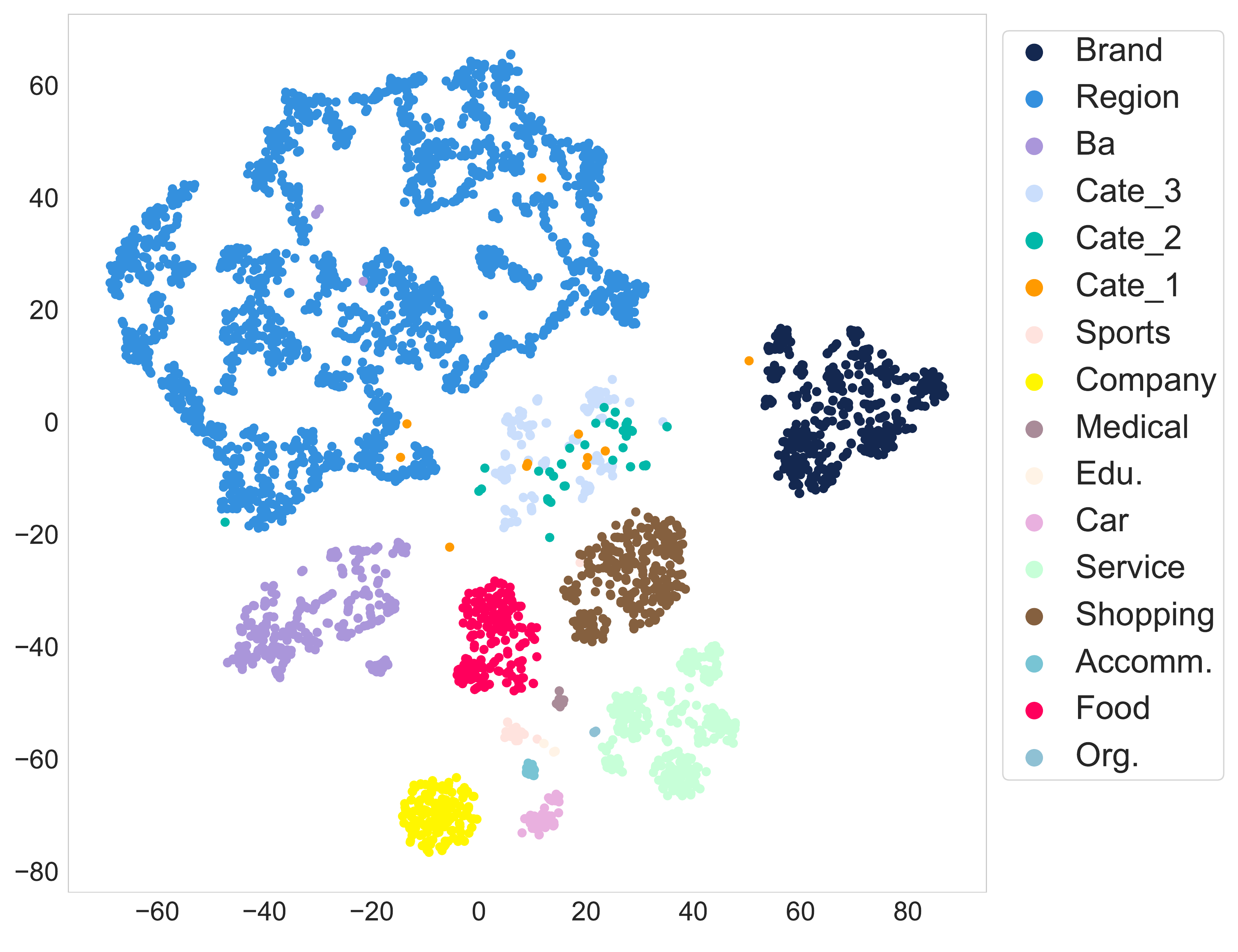}
		\vspace{-10px}
		\caption{t-SNE of pre-trained entity embeddings of shanghai's UrbanKG (better viewed in color).}
		\label{fig:kg_vis_sh}
	\end{figure}

	We further provide the embedding visualization in Shanghai in Figure~\ref{fig:kg_vis_sh}, and a similar clustering  results with Figure~\ref{fig:kg_vis} can also be observed therein. 
	Since the ontology semantics to various entities are quite clear, pre-trained embeddings are successfully separated from different ontology groups. 

\end{document}